\documentclass{article}

\PassOptionsToPackage{numbers, compress}{natbib}

\PassOptionsToPackage{numbers,sort&compress}{natbib}
 \usepackage[main, final]{neurips_2025}
 \usepackage{subcaption}

\setcitestyle{numbers,square,comma}
 \usepackage{wrapfig}
  \usepackage{subcaption}

\usepackage[utf8]{inputenc} % allow utf-8 input
\usepackage[T1]{fontenc}    % use 8-bit T1 fonts
\usepackage{hyperref}       % hyperlinks
\usepackage{url}            % simple URL typesetting
\usepackage{booktabs}       % professional-quality tables
\usepackage{amsfonts}       % blackboard math symbols
\usepackage{nicefrac}       % compact symbols for 1/2, etc.
\usepackage{microtype}      % microtypography
\usepackage{xcolor}         % colors
\usepackage{algorithm}
\usepackage{algpseudocode}
\usepackage{amssymb}
\usepackage{multirow}

\usepackage[table]{xcolor}
\usepackage{booktabs}

\usepackage{booktabs}
\usepackage{multirow}
\usepackage{graphicx}
\usepackage[table]{xcolor}
\title{ROAD-VLA: Robust Online Adaptation via Self-Distillation for Vision-Language-Action Models}

\author{%
  Kejing Wang\thanks{Equal contribution.} \\
  School of Computer Science and Engineering\\
  University of New South Wales\\
  Sydney, Australia \\
  \texttt{kejing.wang@student.unsw.edu.au} \\
  \And
  Toan Nguyen\footnotemark[1] \\
  School of Computer Science and Engineering\\
  University of New South Wales\\
  Sydney, Australia \\
  \texttt{toan.nguyen@unsw.edu.au} \\
  \And
  Minh Hoang Nguyen\footnotemark[1] \\
  Applied Artificial Intelligence Initiative \\ 
  Deakin University \\
  Geelong, Australia \\
  \texttt{s223669184@deakin.edu.au} \\
  \And
  Simon Khan \\
  Air Force Research Laboratory\\
  USA \\
  \texttt{simon.khan@us.af.mil} \\
  \And
  Flora D. Salim\thanks{Corresponding author.} \\
  School of Computer Science and Engineering\\
  University of New South Wales\\
  Sydney, Australia \\
  \texttt{flora.salim@unsw.edu.au} \\
}

\begin{document}

\maketitle

\begin{abstract}
Effective online adaptation of vision-language-action (VLA) models remains challenging, as sparse rewards provide weak supervision for high-dimensional autoregressive action policies. Although self-distillation can in principle provide denser training signals, we find that text-based privileged teachers conditioned on demonstrations, retrieved experiences, or high-level plans are ineffective for VLA adaptation, exposing a modality gap between symbolic guidance and low-level robot actions. We propose \textsc{ROAD-VLA}, an advantage-guided self-distillation framework that constructs a proximal teacher directly in action space by perturbing action-token logits with calibrated advantage estimates. This converts sparse rewards into dense token-level supervision while keeping the teacher close to the current policy. We further derive a policy-improvement lower bound under calibrated advantages and accurate teacher matching. Across seven robotic manipulation environments with in-distribution and out-of-distribution shifts, \textsc{ROAD-VLA} outperforms PPO in nearly all settings, demonstrating robust online VLA adaptation. 

% Our source code is available \href{https://anonymous.4open.science/r/ROAD-1726/}{here}.
\end{abstract}

\section{Introduction}
\label{sec:Intro}

Vision-language-action (VLA) models have emerged as a promising paradigm for
general-purpose robotic manipulation, mapping visual observations and language
instructions directly to low-level actions through large-scale pretraining on diverse
demonstrations~\citep{firoozi2025foundation,ma2026survey}. Recent foundation models such
as OpenVLA~\citep{kim2024openvla} and $\pi_0$~\citep{black2024pi_0} generalize impressively
across tasks and environments. Adapting these pretrained policies to deployment, however,
remains a fundamental challenge: robots routinely face distribution shifts, such as novel
appearances, unseen object configurations, sensor noise, or execution errors, that
pretraining never covered, making effective online adaptation essential for reliable
real-world use.

Reinforcement learning (RL) is a natural framework for such post-training adaptation,
improving VLA policies through interaction without further expert demonstrations. Methods
such as PPO~\citep{schulman2017ppo}, DPO~\citep{rafailov2023direct,zhang2024grape}, and
GRPO~\citep{guo2025deepseek} have been explored for large pretrained policies, and RL can
outperform supervised fine-tuning by exploring beyond the demonstration
support~\citep{liu2026what}. Yet robotic tasks typically provide sparse, delayed rewards,
so standard policy-gradient methods suffer from high variance, optimization instability,
and catastrophic forgetting of pretrained capabilities~\citep{sutton1998reinforcement}.

A compelling alternative is self-distillation~\citep{hinton2015distilling}, which has
recently proven effective for fine-tuning LLMs by using a privileged version of the model
itself as a teacher, converting sparse outcome feedback into dense step-wise
supervision~\citep{agarwal2024policy,zhao2026self,penaloza2026privileged}. This raises an obvious question: \emph{can self-distillation provide a similarly effective adaptation signal for VLA models?} The first thing one might try—conditioning the teacher on demonstrations or high-level text plans—does not work (Section~\ref{sec:ablation_study}). After embodied post-training, current VLA policies retain little reasoning capacity over their LLM backbones~\citep{kim2024openvla}, so purely textual context cannot bridge the gap between language hints and low-level actions~\citep{brohan2022rt1,zhang2024hirt}.

\begin{figure}[t]
    \centering
    \includegraphics[width=1\linewidth]{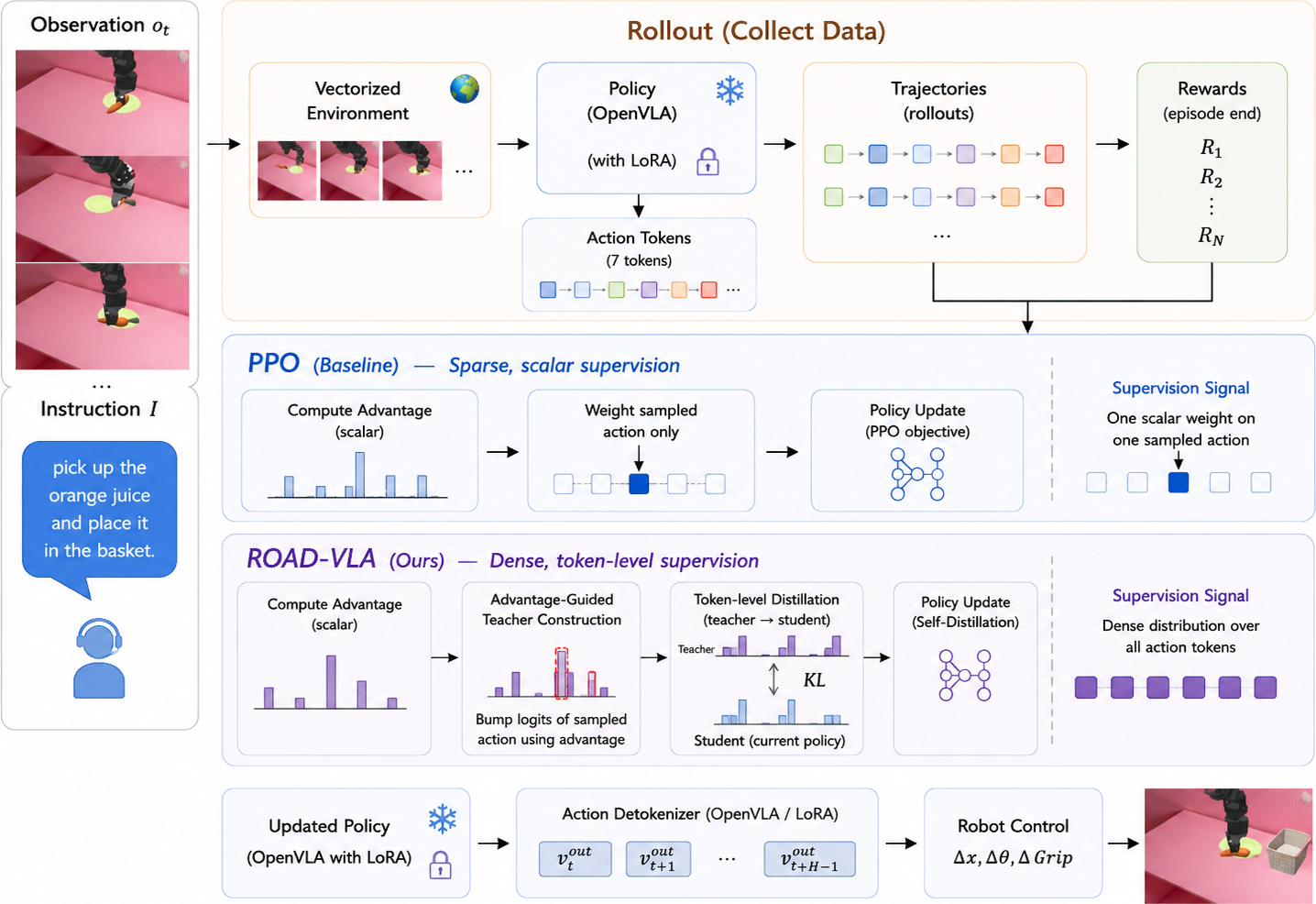}
    \caption{
    Overview of \textsc{ROAD-VLA}. 
    During rollout, OpenVLA collects sparse-reward trajectories, and \textsc{ROAD-VLA} converts advantage estimates into a proximal teacher distribution for dense token-level distillation.
    }
    \label{fig:road_vla_overview}
\end{figure}

We instead propose ROAD-VLA (Figure~\ref{fig:road_vla_overview}), a self-distillation
framework tailored to VLA adaptation. ROAD-VLA builds a stronger teacher directly from the
current policy by perturbing its action logits with advantage estimates, yielding a
proximal teacher that upweights actions estimated to be beneficial. This perturbation is
the closed-form solution of a KL-regularized local improvement problem, and we show it
admits a policy-improvement guarantee under mild calibration conditions. Crucially, the advantage signal, ordinarily a scalar weight on a single sampled action, becomes dense token-level supervision across all action dimensions at every on-policy timestep, addressing the core weakness of PPO without any external teacher or demonstration data at deployment.

We evaluate ROAD-VLA on a comprehensive suite of robot manipulation environments spanning
three axes of distribution shift: visual robustness (unseen backgrounds, dynamic textures,
sensor noise), compositional reasoning (disambiguation among unseen distractors), and
execution robustness (geometry shifts and mid-episode state perturbations). Using
OpenVLA-7B as the base architecture, ROAD-VLA consistently outperforms a strong PPO
baseline across both in-distribution and out-of-distribution settings, confirming that
dense, token-level advantage-guided supervision yields a more generalizable policy than
scalar reward weighting. We summarize our main contributions as follows:
\begin{itemize}
\item We propose ROAD-VLA, a self-distillation framework that constructs a proximal teacher
by perturbing action logits with advantage estimates, converting sparse scalar rewards into
dense token-level supervision without any external teacher model or additional demonstration
data.
\item We provide a formal analysis showing that distilling toward the advantage-guided
proximal teacher admits a policy-improvement guarantee under mild calibration conditions.
\item We evaluate ROAD-VLA on a comprehensive manipulation suite spanning visual,
compositional, and execution shifts, where it consistently outperforms PPO in both in- and
out-of-distribution settings.
\end{itemize}

\section{Related Work}
\label{sec:Related}

\textbf{Vision--Language--Action Models.}
Vision--language--action (VLA) models have become a promising framework for language-conditioned robot control. RT-1 demonstrated that Transformer-based policies can scale with diverse real-world robot data \citep{brohan2022rt1}, while RT-2 connected web-scale vision-language pretraining with robot control by representing actions as tokens \citep{zitkovich2023rt2}. PaLM-E studied embodied multimodal reasoning with continuous sensor inputs \citep{driess2023palme}. More recent open-source robot foundation models, including Octo~\citep{octo2024}, OpenVLA and OpenVLA-OFT~\citep{kim2024openvla,kim2025fine}, $\pi_0$~\citep{black2024pi_0}, and $\pi_{0.5}$~\citep{physicalintelligence2025pi05}, have enabled downstream adaptation of generalist robot policies. We build on OpenVLA~\citep{kim2024openvla}, which represents continuous robot actions as autoregressive discrete action tokens and combines pretrained visual and language components \citep{zhai2023sigmoid,karamcheti2024prismatic,oquab2024dinov2}. Despite strong pretrained capabilities, current VLA policies remain brittle under visual, compositional, and execution-level distribution shifts, motivating robust online adaptation. Existing approaches primarily focus on scaling pretraining data, model architectures, or fine-tuning procedures, whereas we study how to adapt VLA policies online from sparse reward feedback. In particular, we introduce an advantage-guided self-distillation framework that converts sparse rewards into dense token-level supervision by constructing a proximal teacher directly in the action-token space.

\textbf{Online Adaptation and Reinforcement Learning.}
Online reinforcement learning provides a natural way to adapt VLA policies through environment interaction when expert demonstrations are limited. PPO is a widely used policy optimization method \citep{schulman2017ppo}, and recent work shows that RL can improve VLA generalization, with PPO serving as a strong baseline for VLA adaptation \citep{liu2026what,liu2026what}. However, PPO-style updates use advantage estimates as scalar weights on sampled actions, which provides sparse and high-variance supervision for high-dimensional autoregressive action tokens. Recent work has studied more stable online or test-time adaptation for VLAs. RobustVLA improves VLA robustness through reinforcement post-training with robustness-aware regularization \citep{zhang2025robustvla}. TT-VLA studies test-time reinforcement learning for on-the-fly VLA adaptation using dense task-progress feedback \citep{liu2026onthefly}. In contrast, ROAD-VLA converts advantage estimates into dense token-level teacher distributions, stabilizing online adaptation without relying solely on scalar policy-gradient updates.

\textbf{Distillation for Policy Adaptation.}
Knowledge distillation trains a student to match soft teacher distributions \citep{hinton2015distilling}, and policy distillation extends this idea to reinforcement learning policies \citep{rusu2015policy}. Distillation has also been used in continual learning to preserve previous capabilities while adapting to new tasks \citep{li2016learning}. Recent self-distillation methods for large language models further show that on-policy distillation can provide denser supervision than sparse outcome feedback \citep{shenfeld2026self,zhao2026self}. Distillation for VLA models is closely related to our work but differs in objective and teacher construction. ActDistill~\citep{ye2025actdistill} uses action-guided self-derived distillation to train efficient VLA models, mainly targeting model compression and inference efficiency. VLA-OPD~\citep{zhong2026vlaopd} bridges offline SFT and online RL through on-policy distillation with an expert teacher that provides dense token-level supervision. ROAD-VLA instead constructs a proximal teacher from the current policy itself by perturbing action-token logits according to calibrated advantage estimates. This avoids requiring an external expert teacher and turns scalar RL feedback into dense supervision over all action tokens.

\section{Background}
\label{sec:Background}

\subsection{Vision--Language--Action models}

We study a pretrained vision--language--action (VLA) model, OpenVLA~\citep{karamcheti2024prismatic}, that maps an image observation at time step $t$, denoted by $o_t \in \mathcal{O}$, together with a language instruction $l \in \mathcal{L}$ to a continuous action $a_t \in \mathbb{R}^{d_a}$ via autoregressive next-token prediction. The visual input $o_t$ is encoded by a fused SigLIP~\citep{zhai2023sigmoid}--DINOv2~\citep{oquab2024dinov2} encoder and projected into language embedding space, while $l$ is tokenized using the Llama 2 tokenizer~\citep{touvron2023llama}. The resulting multimodal token sequence is then processed by a causal transformer decoder. To enable next-token prediction, each dimension of $a_t$ is discretized into one of $256$ bins, and the decoder predicts the corresponding discrete action tokens.

\subsection{Problem Formulation}

We consider the problem of \emph{online} adaptation of a pretrained VLA policy $\pi_{\theta_0}$ to a target environment $\mathcal{M}^{\mathrm{tar}}$, while preserving the general capabilities acquired during pretraining.  Let $\mathcal{M}^{\mathrm{tar}} = (\mathcal{S}, \mathcal{A}, \mathcal{P}, \mathcal{R}, \mathcal{O}, \mathcal{L}, p(s_0), \gamma)$ denote a language-conditioned partially observable Markov decision process (POMDP), where $\mathcal{S}$, $\mathcal{A}$, $\mathcal{O}$, and $\mathcal{L}$ are the state, action, observation, and language spaces, respectively, $\mathcal{P}$ is the transition kernel, $\mathcal{R}$ is the reward function, $p(s_0)$ is the initial-state distribution, and $\gamma \in [0,1)$ is the discount factor. At the start of each episode, an instruction $l \in \mathcal{L}$ and an initial state $s_0 \sim p(s_0)$ are sampled. At time step $t$, the policy conditions on the instruction $l$ and the recent observation history $o_{t-H+1:t}$ to produce an action
\[
a_t \sim \pi_\theta(\cdot \mid o_{t-H+1:t}, l),
\]
thereby inducing a trajectory $\tau = (o_0, a_0, o_1, a_1, \dots)$ under the environment dynamics. Formally, starting from a pretrained policy $\pi_{\theta_0}$, we seek adapted parameters $\theta$ that improve performance in the target environment. A common adaptation strategy is supervised fine-tuning (SFT), which minimizes the standard imitation loss
\begin{equation}
\theta^\star
=
\arg\min_\theta \mathcal{L}_{\mathrm{SFT}}(\theta),
\qquad
\mathcal{L}_{\mathrm{SFT}}(\theta)
=
-\mathbb{E}_{(o_{t-H+1:t},\,l,\,a_t)\sim \mathcal{D}^{\mathrm{tar}}}
\big[
\log \pi_\theta(a_t \mid o_{t-H+1:t}, l)
\big].
\end{equation}
In practice, obtaining expert-labelled trajectories in the target environment can be costly, so a common alternative is to fine-tune the policy on trajectories generated online by a teacher policy. For direct online RL fine-tuning, one often optimizes a policy-gradient objective in the PPO form:
\begin{equation}
\label{eq:ppo_finetuning}
\mathcal{L}_{\mathrm{PPO}}(\theta)
=
-
\mathbb{E}_t\!\left[
\min\!\Big(
r_t(\theta)\hat{A}_t,\;
\mathrm{clip}\!\big(r_t(\theta),1-\epsilon,1+\epsilon\big)\hat{A}_t
\Big)
\right],
\end{equation}
where the likelihood ratio is defined as
\begin{equation}
r_t(\theta)
=
\frac{\pi_\theta(a_t \mid o_{t-H+1:t}, l)}
{\pi_{\theta_0}(a_t \mid o_{t-H+1:t}, l)}.
\end{equation}
Here, $\hat{A}_t$ denotes the advantage estimate and $\epsilon>0$ is the clipping threshold. Related objectives, including GRPO~\citep{shao2024deepseekmath} and DPO~\citep{rafailov2023direct}, can also be applied when relative feedback or preference data are available. However, recent evidence suggests that PPO remains highly competitive, and often performs best, for VLA adaptation~\citep{liu2026what}. Despite their appeal, these objectives share a fundamental limitation: their effectiveness is bounded by the quality of the adaptation signal, whether obtained from expert demonstrations, teacher-generated rollouts, or reward feedback from environment interaction. In VLA settings, such signals are often sparse, noisy, suboptimal, or misaligned with the target task, causing the adapted policy to inherit these imperfections and become brittle under distribution shift. This motivates a central question: \emph{how can we construct a stronger teacher that provides denser, more reliable supervision for robust RL fine-tuning in new environments?}

\section{Method}
\label{sec:Method}

\subsection{Self-Distillation}

To improve the adaptation signal in VLA post-training, we adopt a self-distillation strategy inspired by its recent success in fine-tuning LLMs~\citep{zhao2026self,shenfeld2026self}. Rather than relying on a larger external teacher, we induce a stronger teacher from the same-sized VLA by conditioning it on privileged training-time information ($\mathcal I$), such as demonstrations, trajectory context, or outcome annotations. The deployable student first collects on-policy rollouts using only test-time inputs. We then re-query the same VLA at each visited state with privileged information ($\mathcal{I}$) to obtain an advantage-aligned teacher distribution, which the student distills without access to ($\mathcal{I}$). 

This yields an \emph{advantage-aligned proximal} teacher at each visited state, so the student learns from a privileged refinement of its own policy rather than from a separate, potentially misaligned oracle. The key effect is to densify supervision: sparse trajectory-level feedback is converted into learning signals across decision steps, autoregressive action tokens, and privileged reward components (see Theorem~1). Given an on-policy student rollout

\[
\hat{\tau}
=
\bigl((o_1,\hat{a}_1),\dots,(o_T,\hat{a}_T)\bigr)
\sim
\pi_S(\cdot\mid l),
\]
we distill the privileged teacher along the student’s own trajectory. Since each OpenVLA action is represented as \(K=7\) discrete tokens, we apply token-level distillation uniformly across all action dimensions. At step \(t\), let
\begin{equation}
p_{t,k}^{T}
=
\pi_T^{(k)}(\cdot\mid o_{\le t},\mathcal{I},\hat{a}_{<t}),
\qquad
p_{t,k}^{S}
=
\pi_S^{(k)}(\cdot\mid o_{\le t},l,\hat{a}_{<t})
\end{equation}
denote the teacher and student distributions for the \(k\)-th action token, respectively. We minimize the token-level Jensen--Shannon distillation objective
\begin{equation}
\mathcal{L}_{\mathrm{SD}}
=
\mathbb{E}_{\hat{\tau}\sim \pi_S(\cdot \mid l)}
\left[
\frac{1}{TK}
\sum_{t=1}^{T}
\sum_{k=1}^{K}
\mathrm{JSD}\!\left(
p_{t,k}^{T},
p_{t,k}^{S}
\right)
\right],
\end{equation}
where
\begin{equation}
\mathrm{JSD}(p^T,p^S)
=
\frac{1}{2}\mathrm{KL}(p^T\|m)
+
\frac{1}{2}\mathrm{KL}(p^S\|m),
\qquad
m=\frac{1}{2}(p^T+p^S).
\end{equation}

Although JSD is a natural symmetric matching loss, we use forward KL
\(\mathrm{KL}(p^T\|p^S)\) as the default objective. Since the privileged teacher
defines the local improvement target, the student should cover the probability
mass assigned by this improved distribution; this is exactly the direction
controlled by the distillation penalty in
Theorem~1. Empirically, we find that forward KL gives
more stable training and stronger adaptation. The remaining question is how to
construct a reliable self-privileged signal \(\mathcal I\) for VLA adaptation.

\subsection{Design of Privileged Information}
\label{sec:pi_design}

\subsubsection{Text-guided $\mathcal{I}$}
\label{sec:Text_pi}
% A key prerequisite for the privileged context \(\mathcal{I}\) is that conditioning on it should improve the current policy, so that the resulting privileged policy can act as a stronger teacher. A natural design, inspired by self-distillation in LLM fine-tuning, is to use \(\mathcal{I}\) as a text-based in-context hint that guides the teacher towards better actions.

% \begin{center}
% \fbox{
% \begin{minipage}{0.92\linewidth}
% \small
% \textbf{Text-guided \(\mathcal{I}\).}
% \emph{Reference plan for instruction ``\(\{l\}\)'':}
% Approach the object \(\rightarrow\) align for grasping \(\rightarrow\) grasp and lift
% \(\rightarrow\) move to the target \(\rightarrow\) align for placement \(\rightarrow\) place and retract.
% \end{minipage}
% }
% \end{center}

% However, we find that this intuitive design does not reliably improve VLA adaptation. Purely text-based hints, such as high-level plans or subgoal descriptions, often fail to provide grounded action-level supervision. We hypothesize two reasons: first, VLA post-training on instruction-action pairs may weaken the LLM backbone’s general reasoning ability; second, adding more text does not necessarily translate into better continuous control due to the modality gap between language hints and low-level actions.

A natural way to instantiate the privileged context \(\mathcal I\) is to provide
the teacher with text-based in-context hints. The goal is to make the privileged
policy stronger than the deployable student by giving it additional guidance
during training. We consider three variants, with representative templates shown below.

\textbf{MCTS PI} provides local action-level hints by retrieving the most
similar successful transition from a growing tree of past successful rollouts
and verbalising the retrieved action:
\begin{center}
\fbox{
\begin{minipage}{0.92\linewidth}
\small
\textbf{MCTS PI.}
\emph{Hint (q=1.20): move (+0.05, \(-\)0.02, +0.10), rotate (+0.00, +0.05, +0.00), gripper open.}
\end{minipage}
}
\end{center}

\textbf{RelSpatial PI} instead provides real-time egocentric spatial
descriptions from simulator proprioception, describing the relative positions of
the object, target, and gripper:
\begin{center}
\fbox{
\begin{minipage}{0.92\linewidth}
\small
\textbf{RelSpatial PI.}
\emph{carrot is right, ahead, below gripper, not grasped. plate is ahead, left, below gripper. carrot to plate: move ahead and left. gripper open.}
\end{minipage}
}
\end{center}

\textbf{Plan+RelSpatial PI} combines a static high-level task plan with the
current spatial state, giving the teacher both task structure and local
grounding:
\begin{center}
\fbox{
\begin{minipage}{0.92\linewidth}
\small
\textbf{Plan+RelSpatial PI.}
\emph{Reference plan for ``put carrot on plate'': 1.~Approach the object.
2.~Align for grasping. 3.~Grasp and lift. 4.~Move toward the target.
5.~Align for placement. 6.~Place and retract. Current state: carrot is right,
ahead, below gripper, not grasped. plate is ahead, left, below gripper.
carrot to plate: move ahead and left. gripper open.}
\end{minipage}
}
\end{center}

Empirically, however, these text-guided privileged contexts do not reliably
improve VLA adaptation. Even when grounded in successful motor experience or
real-time spatial state, language hints provide weak action-level supervision.
We attribute this to two factors. First, VLA post-training on
instruction-action pairs may reduce the model's ability to exploit general
in-context language hints. Second, text descriptions remain separated from the
low-level action space: the policy must still translate verbal guidance into
precise continuous control. These observations motivate replacing
language-based privileged context with an action-centric, value-based signal.

\subsubsection{Advantage-guided privileged context}
\label{sec:adv_guided_context}

\paragraph{From outcome reward to a local teacher.}
During VLA fine-tuning, the operative question is not \emph{which} instruction to
follow, but \emph{which of the policy's own sampled actions deserve reinforcement}.
The advantage estimate answers exactly this: a positive advantage marks an action that
outperforms the critic baseline, a negative one marks below-baseline behavior. PPO uses
this scalar only to reweight the likelihood of the sampled action
(Eq.~\ref{eq:ppo_finetuning}). We instead use it to construct a \emph{local teacher
distribution} over action tokens, converting sparse outcome-level feedback into dense
token-level supervision on the student's own on-policy states. We thus instantiate the
privileged context $\mathcal I$ in this action-centric form, and write the conditioning
context at step $t$ as $h_t=(o_{\le t},\,l,\,\hat a_{<t})$ for brevity.

\paragraph{Calibrated, agreement-gated advantage.}
For each sampled action $\hat a_t$, we form a step-wise advantage signal. To reduce
variance we blend the intrinsic estimate $\hat A_t^{\mathrm{int}}$ from the current policy
with a reference estimate $\hat A_t^{\mathrm{ref}}$ from a frozen PPO critic. As the two
estimates live on different scales, we first match the reference to the batch statistics
of the intrinsic signal,
\begin{equation}
\tilde A_t^{\mathrm{ref}}
=
\mu_{\mathrm{int}}
+
\frac{\sigma_{\mathrm{int}}}{\sigma_{\mathrm{ref}}+\varepsilon}
\bigl(\hat A_t^{\mathrm{ref}}-\mu_{\mathrm{ref}}\bigr),
\end{equation}
and combine them only when they agree on the sign of the advantage:
\begin{equation}
\hat A_t^{\mathrm{mix}}
=
\hat A_t^{\mathrm{int}}
+
\alpha\,g_t\bigl(\tilde A_t^{\mathrm{ref}}-\hat A_t^{\mathrm{int}}\bigr),
\qquad
g_t=\mathbf 1\!\left[
\mathrm{sign}(\hat A_t^{\mathrm{int}})
=
\mathrm{sign}(\tilde A_t^{\mathrm{ref}})
\right].
\end{equation}
By default we set $\alpha=0.5$, weighting the intrinsic and reference advantages equally.
The reference critic therefore acts as a sign-consistent correction to the intrinsic
estimate, and is discarded whenever the two disagree, preventing a miscalibrated critic
from overriding on-policy evidence.

\paragraph{Perturbation weight.}
We then standardize and clip the mixed advantage to obtain the weight
\begin{equation}
\omega_t
=
\mathrm{clip}\!\left(
\frac{\hat A_t^{\mathrm{mix}}-\mu_{\mathrm{mix}}}{\sigma_{\mathrm{mix}}+\varepsilon},
\,-c,\,c
\right).
\end{equation}
One might additionally apply a ReLU $[\omega_t]_+$ to retain only above-baseline signals,
on the intuition that negative advantage estimates are noisier and best left alone. In
practice we find this unnecessary (Appendix~\ref{subsec:results_adv_sign_pos_gate}): the
signed weight does just as well, since below-baseline tokens still carry useful learning
signal. We keep the signed $\omega_t$ throughout, and fix the clipping to $c=2.0$ without
tuning.

\paragraph{Advantage-guided teacher.}
The teacher is simply a logit-perturbed copy of the student. Writing
$z_{t,k}(u)$ for the student logit of value $u\in\mathcal V_k$ at position $k$ and
$p^\theta_{t,k}=\mathrm{softmax}(z_{t,k})$ for the student token distribution, we define
\begin{equation}
q^\star_{t,k}
=
\mathrm{softmax}\!\left(
z_{t,k}+\eta\,\omega_t\,e_{\hat a_{t,k}}
\right),
\end{equation}
where $e_{\hat a_{t,k}}$ is the one-hot indicator of the sampled token and $\eta>0$ sets
the perturbation strength, fixed to $\eta=1.0$ without tuning. This makes the teacher a
\emph{locally improved} student: it nudges the probability of each sampled token up or
down according to the sign of $\omega_t$, by an amount that reflects how useful the
rollout evidence deems that action.

\paragraph{Proximal interpretation.}
The logit perturbation is not ad-hoc: at each token position it is the closed-form
solution of a KL-regularized improvement problem,
\begin{equation}
q_{t,k}^\star
=
\mathrm{arg\,max}_{q\in\Delta(\mathcal V_k)}
\left\{
\mathbb E_{u\sim q}\bigl[r_{t,k}(u)\bigr]
-
\tau\,\mathrm{KL}\!\bigl(q\,\|\,p^\theta_{t,k}(\cdot\mid h_t)\bigr)
\right\},
\end{equation}
whose optimum is the exponential tilt
$q_{t,k}^\star(u)\propto p^\theta_{t,k}(u)\exp(r_{t,k}(u)/\tau)$. With the one-point token
reward $r_{t,k}(u)=\tau\eta\,\omega_t\,\mathbf 1[u=\hat a_{t,k}]$, this recovers exactly the
logit shift above; setting $\tau=1$ gives $\eta=1/\tau=1$, and the reward simplifies to
$r_{t,k}(u)=\omega_t\,\mathbf 1[u=\hat a_{t,k}]$. The full-action reward is the token sum
$r_t(a)=\sum_{k}r_{t,k}(a_k)$, and the teacher composes autoregressively,
$q_t^\star=\prod_{k}q_{t,k}^\star$, the KL term holding each token teacher within a trust
region of the corresponding student conditional. The full derivation is deferred to
Appendix~\ref{app:prox_teacher_derivation}.

\paragraph{Distillation objective.}
We then distill the deployable student onto this local teacher with a token-level
teacher-to-student KL,
\begin{equation}
\mathcal L_{\mathrm{AGD}}
=
\mathbb E_{\hat\tau}
\left[
\frac{1}{TK}
\sum_{t=1}^{T}\sum_{k=1}^{K}
\mathrm{KL}\!\bigl(q^\star_{t,k}\,\|\,p^\theta_{t,k}\bigr)
\right].
\end{equation}
Where PPO collapses the advantage into a single scalar multiplying the sampled-action
likelihood, ROAD-VLA expands it into a full teacher distribution over action tokens,
giving dense supervision at every on-policy step while the KL-proximal construction
holds that teacher close to the policy. On-policy rollouts, calibrated advantage mixing,
critic-agreement gating, signed logit perturbation, and teacher-to-student KL
distillation together form the ROAD-VLA adaptation objective, summarized in
Algorithm~\ref{alg:roadvla}.

\subsection{Theoretical results}
\label{sec:theoretical_results}

\paragraph{Theorem 1 (Self-privileged advantage distillation).}
Consider a finite-horizon VLA policy \(\pi_\theta\) with horizon \(T\). Let
\(q_t^\star=\prod_{k=1}^{K}q_{t,k}^\star\) be the per-token advantage-guided teacher of
Sec.~\ref{sec:adv_guided_context}, with full-action reward
\(r_t(a)=\sum_{k=1}^{K}r_{t,k}(a_k)\). Since the token rewards may be signed, center each one under the corresponding student
conditional and sum,
\begin{equation}
\bar r_t(a)
=
\sum_{k=1}^{K}
\Bigl(
r_{t,k}(a_k)
-
\mathbb E_{v\sim p^\theta_{t,k}(\cdot\mid a_{<k},h_t)}\!\bigl[r_{t,k}(v)\bigr]
\Bigr).
\end{equation}
Each per-token subtraction is independent of the token value and leaves the per-token
teacher unchanged. Suppose that, for policy contexts \(h_t\)
in the support of \(d_t^{\pi_{\theta'}}\), the centered reward is aligned with the true
advantage under the teacher:
\begin{equation}
\mathbb E_{a\sim q_t^\star(\cdot\mid h_t)}
\left[
A_t^{\pi_\theta}(h_t,a)
\right]
\ge
\beta
\mathbb E_{a\sim q_t^\star(\cdot\mid h_t)}
\left[
\bar r_t(a)
\right]
-
\epsilon_{\mathrm{cal}},
\qquad
\beta>0.
\end{equation}
Assume also that \(|A_t^{\pi_\theta}(h_t,a)|\le B\). Let \(\pi_{\theta'}\) be the distilled
student, and define the teacher-to-student distillation error at \(h_t\) as
\begin{equation}
D_t^{\mathrm{dist}}(h_t)
=
\mathrm{KL}
\left(
q_t^\star(\cdot\mid h_t)
\,\middle\|\,
\pi_{\theta'}(\cdot\mid h_t)
\right).
\end{equation}
Then, with \(C=\sqrt2\) (from Pinsker's inequality),
\begin{equation}
J(\pi_{\theta'})
\ge
J(\pi_\theta)
+
\frac{1}{T}
\sum_{t=1}^{T}
\mathbb E_{h_t\sim d_t^{\pi_{\theta'}}}
\left[
\beta\tau
\mathrm{KL}
\left(
q_t^\star(\cdot\mid h_t)
\,\middle\|\,
\pi_\theta(\cdot\mid h_t)
\right)
-
CB
\sqrt{D_t^{\mathrm{dist}}(h_t)}
\right]
-
\epsilon_{\mathrm{cal}} .
\end{equation}
Moreover, since both \(q_t^\star\) and \(\pi_\theta\) factorize
autoregressively, the teacher-policy KL decomposes by the chain rule as
\begin{equation}
\mathrm{KL}
\left(
q_t^\star(\cdot\mid h_t)
\,\middle\|\,
\pi_\theta(\cdot\mid h_t)
\right)
=
\sum_{k=1}^{K}
\mathbb E_{a_{<k}\sim q_t^\star}
\left[
\mathrm{KL}
\left(
q_{t,k}^\star(\cdot\mid a_{<k},h_t)
\,\middle\|\,
p^\theta_{t,k}(\cdot\mid a_{<k},h_t)
\right)
\right].
\end{equation}

\paragraph{Intuition.}
Centering subtracts each token reward's mean under the corresponding student conditional,
so it vanishes in expectation under the student at every position while leaving the
exponential-tilted teacher unchanged. The calibration condition relates the teacher's
shaped reward to true advantage, and per-token proximal optimality (summed over tokens via
the chain rule) gives
\[
\mathbb E_{q_t^\star}[\bar r_t]
\ge
\tau\,
\mathrm{KL}\!\left(q_t^\star(\cdot\mid h_t)\,\middle\|\,\pi_\theta(\cdot\mid h_t)\right).
\]
Thus the teacher-policy KL becomes the improvement term in the bound, while the student's
mismatch to the teacher appears as the distillation cost. The autoregressive KL
decomposition connects this improvement term to the token-level supervision optimized by
\(\mathcal L_{\mathrm{AGD}}\).

\section{Experiments}
\label{sec:Experiments}

We design our evaluation to test the core hypothesis of ROAD-VLA: that advantage-weighted online self-distillation provides a superior regularizer for VLA adaptation compared to standard RL, particularly under distribution shift.

\subsection{Experimental Setup and Baselines}

\textbf{Testing Environments.} To probe the boundaries of VLA robustness, we utilize a suite of environments categorized by three distinct axes of distribution shift (detailed in Appendix~\ref{app:env_details}). \textbf{Visual Robustness:} Includes \texttt{VR-UnseenTable} (background shift), \texttt{VR-DynamicTexture} (non-stationary visual noise), and \texttt{VR-DynamicNoise} (sensor perturbations). \textbf{Compositional Reasoning:} Evaluates generalization to \texttt{CR-MultiObject} and \texttt{CR-MultiReceptacle}, requiring the model to disambiguate targets from unseen distractors. \textbf{Execution Robustness:} Tests temporal adaptation via \texttt{ER-InitPose} (geometry shift) and \texttt{ER-Repositioning} (mid-episode state perturbation).

\textbf{Implementation and Baselines.} We use \texttt{OpenVLA-7B}~\citep{kim2024openvla} as our base architecture. We evaluate our method against a strong PPO baseline, which represents the standard approach for on-policy RL adaptation in Vision-Language-Action models. Both ROAD-VLA and PPO are initialized from a shared foundation warm-up checkpoint—fine-tuned on 140 expert trajectories—to ensure a consistent baseline for perception and low-level control~\citep{liu2026what}. To ensure a controlled comparison, both methods share identical rollout buffers, reward structures, and optimization hyperparameters (details in Appendix.~\ref{app:imp_details}.

\textbf{Experimental Scope.} With this setup in place, we structure our analysis around four concrete questions: \textbf{(Q1)} Does \textsc{ROAD-VLA} achieve higher task success than PPO across ID and OOD settings (Section~\ref{sec:benchmark_results})? \textbf{(Q2)} Does \textsc{ROAD-VLA} adapt more efficiently during online interaction in terms of convergence speed, peak performance, and late-stage stability (Section~\ref{sec:adaptation_dynamics})? \textbf{(Q3)} Why does advantage-guided distillation work, and what do policy entropy, advantage weight evolution, and critic agreement reveal about its robustness gains (Section~\ref{sec:why_distillation})? \textbf{(Q4)} How does each component contribute, and what do ablations over the privileged information design, distillation loss, mixing coefficient, and agreement gate reveal (Section~\ref{sec:ablation_study})? Finally, we complement these quantitative analyses with a qualitative failure mode analysis in Appendix~\ref{sec:failure_mode_analysis}, examining how \textsc{ROAD-VLA} alters the dominant failure patterns under representative OOD conditions.

\subsection{Benchmark Results: Robustness to Distribution Shift}
\label{sec:benchmark_results}
\begin{table*}[t]
\centering
\tiny
\setlength{\tabcolsep}{4pt}
\renewcommand{\arraystretch}{1.}
\caption{Comparison of task success rates (\%) between ROAD-VLA and PPO across evaluation environments. All results are reported as mean $\pm$ standard deviation over 3 random seeds; the superior result in each row is bolded. $\Delta$ represents the performance degradation from In-Distribution (ID) to Out-of-Distribution (OOD), calculated as $\Delta = \mathrm{ID} - \mathrm{OOD}$ (in \%). The smaller the $\Delta$ the better. }
\label{tab:roadvla_results}

\resizebox{\textwidth}{!}{%
\begin{tabular}{p{2cm} p{3cm} cc cc cc}
\toprule
\textbf{Category} & \textbf{Environment Name} 
& \multicolumn{2}{c}{\textbf{ID}} 
& \multicolumn{2}{c}{\textbf{OOD}}
& \multicolumn{2}{c}{\textbf{$\Delta$}} \\
\cmidrule(lr){3-4} \cmidrule(lr){5-6} \cmidrule(lr){7-8}
& 
& \textbf{PPO} & \textbf{ROAD-VLA} 
& \textbf{PPO} & \textbf{ROAD-VLA}
& \textbf{PPO} & \textbf{ROAD-VLA} \\
\midrule

\textbf{Visual Robustness} 
& \texttt{VR-UnseenTable} & 88 ± 3 & \textbf{93 ± 2} & 87 ± 4 & \textbf{92 ± 1} & 1 & 1 \\
& \texttt{VR-DynamicTexture} 
& 87 $\pm$ 1 
& \textbf{88 $\pm$ 1} 
& 65 $\pm$ 5   
& \textbf{69 $\pm$ 5} 
& 22 & \textbf{19} \\
& \texttt{VR-DynamicNoise} 
& 85 ± 4 & \textbf{90 ± 2} & 66 ± 3 & \textbf{70 ± 2} & \textbf{19} & 20 \\

\midrule

\textbf{Compositional Reasoning} 
& \texttt{CR-MultiObject} & 78 ± 6 & \textbf{80 ± 6} & 61 ± 6 & \textbf{63 ± 4} & 17 & 17 \\
& \texttt{CR-MultiReceptacle} & 84 ± 3 & 84 ± 3 & 57 ± 1 & \textbf{62 ± 2} & 27 & \textbf{22} \\

\midrule

\textbf{Execution Robustness} 
& \texttt{ER-InitPose} 
& 87 $\pm$ 0 
& \textbf{91 $\pm$ 1} 
& 75 $\pm$ 8  
& \textbf{79 $\pm$ 7} 
& 12 & 12 \\

& \texttt{ER-Repositioning} 
& \textbf{89 $\pm$ 3} 
& 88 $\pm$ 0 
& 73 $\pm$ 3 
& \textbf{77 $\pm$ 5} 
& 16 & \textbf{11} \\

\midrule
\midrule
\textbf{Average} 
& All 
& 85 $\pm$ 3 
& \textbf{88 $\pm$ 2} 
& 69 $\pm$ 4 
& \textbf{73 $\pm$ 4} 
& 16.3 & \textbf{14.6} \\

\bottomrule
\end{tabular}%
}
\end{table*}

We evaluate ROAD-VLA under both in-distribution (ID) and out-of-distribution (OOD) settings to measure task performance and robustness to distribution shift. As shown in Table~\ref{tab:roadvla_results}, ROAD-VLA consistently outperforms PPO across most environments, achieving higher success rates with, on average, lower degradation margins ($\Delta$). Averaged across all environments, ROAD-VLA improves ID performance from 85\% to 88\% and OOD performance from 69\% to 73\%, while reducing the average degradation from 16.3\% to 14.6\%. ROAD-VLA shows particularly strong robustness under visual and execution-level shifts. In \texttt{VR-DynamicTexture}, the degradation margin is reduced from 22\% to 19\%, while in \texttt{ER-Repositioning} it decreases from 16\% to 11\%. Additionally, ROAD-VLA improves ID success rates in environments such as \texttt{ER-InitPose} and \texttt{VR-UnseenTable}. These findings suggest that ROAD-VLA learns a more stable and transferable policy, making it a promising framework for deployment in dynamically changing environments.

% \begin{figure}[t]
%     \centering
%     \includegraphics[width=1\linewidth]{source/figures/results_combined.pdf}
%     \caption{(a) Online adaptation trajectories under OOD conditions. ROAD-VLA converges faster, attains higher peak success, and exhibits stronger late-stage stability than PPO. Shaded regions indicate performance variability. (b) OOD grasp success rate on \texttt{VR-DynamicTexture}. ROAD-VLA reaches its peak grasp rate 15 steps earlier than PPO and maintains a consistent advantage throughout training. (c) Policy entropy over training on \texttt{ER-Repositioning}, showing sustained exploration compared to PPO.}
%     \label{fig:finding_1}
% \end{figure}

\subsection{Adaptation Dynamics Under Online Interaction.}
\label{sec:adaptation_dynamics}

\begin{figure}[t]
    \centering
    \includegraphics[width=1\linewidth]{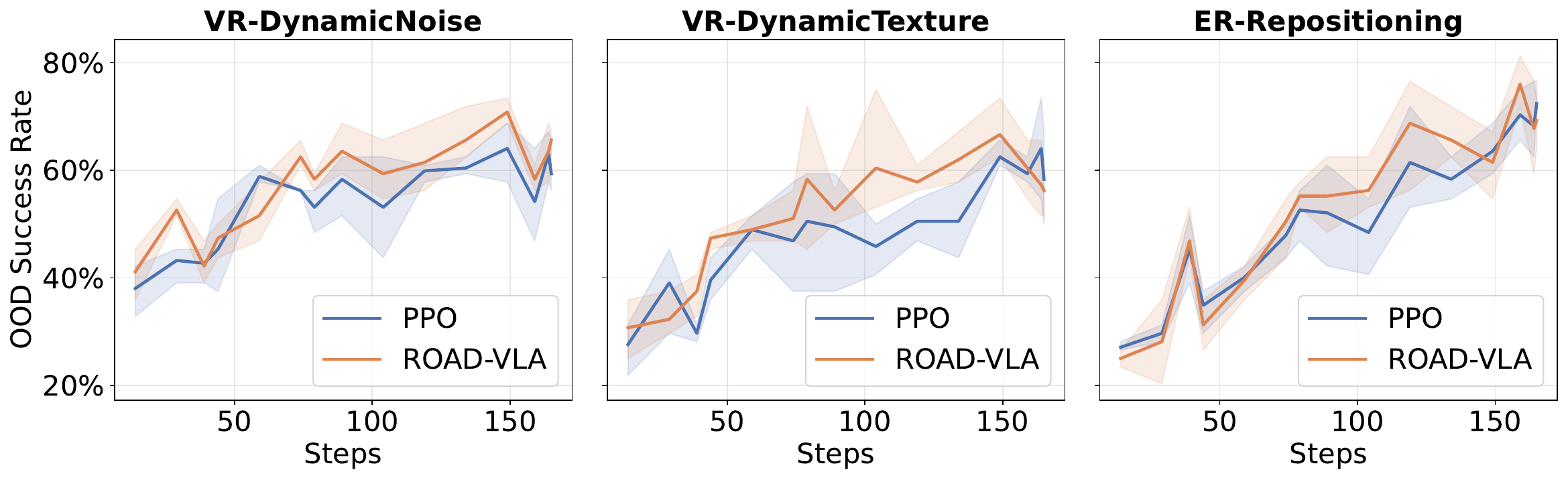}
    \caption{ Online adaptation trajectories under OOD conditions. ROAD-VLA converges faster, attains higher peak success, and exhibits stronger late-stage stability than PPO across all environments. Shaded regions indicate performance variability during training.}
    \label{fig:adaptation_dynamics}
\end{figure}

To evaluate adaptation efficiency, we analyze performance trajectories during online interaction (Figure~\ref{fig:adaptation_dynamics}). Starting from baseline OOD success rates of 27--31\%, ROAD-VLA demonstrates superior sample efficiency, consistently leading PPO throughout the mid-training phase across all environments. In \texttt{VR-DynamicNoise}, ROAD-VLA maintains a dominant lead, peaking at 70.8\% (vs. 64.1\% for PPO) and concluding with a consistent 4\% advantage. Similarly, in \texttt{ER-Repositioning}, our framework reaches a higher mid-training peak of 76.0\% at step 159, outperforming PPO by 8 points during this critical window. This acceleration likely stems from our distillation objective; while PPO relies on sparse terminal rewards, our advantage-weighted signal provides dense, local gradients that utilize the reference model as a structured exploration prior. Beyond speed, ROAD-VLA achieves higher peak OOD success with improved stability. During the final 30\% of training, our method exhibits lower variance than PPO (4\% vs. 6\% on \texttt{VR-DynamicTexture}). We hypothesize that this stability is driven by the agreement gate mechanism, which filters distillation gradients when online and reference critics disagree. By preventing the injection of conflicting signals, the gate regularizes the adaptation process against the high-entropy noise characteristic of visual and execution perturbations.

% \begin{wrapfigure}{r}{0.45\textwidth}
%   \centering
%     \includegraphics[width=1\linewidth]{source/figures/adaptation_analysis.pdf}
%     \caption{
%     Online adaptation trajectories under OOD conditions. ROAD-VLA converges faster, attains higher peak success, and exhibits stronger late-stage stability than PPO across both \texttt{ER-Repositioning} and \texttt{VR-DynamicTexture}. Shaded regions indicate performance variability during training.
%     }
%     \label{fig:adaptation_dynamics}
% \end{wrapfigure}

\subsection{Why Distillation Works?}
\label{sec:why_distillation}
% We present why $A_t$ provides a good insight to better teacher?, distribution of logits?

% Signal of training along the trajectory, better than SFT due to "dense step-wise supervision beyond sparse outcome-level rewards".

\begin{figure}[t]
    \centering
    \includegraphics[width=1\linewidth]{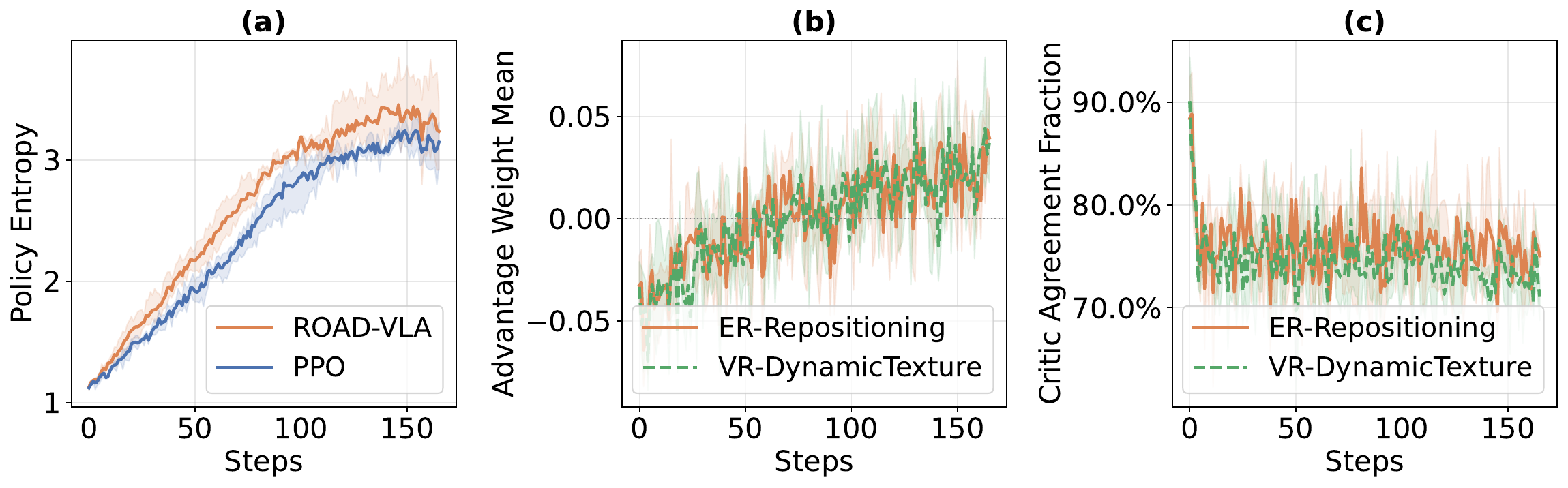}
    \caption{(a) Policy entropy over training on \texttt{ER-Repositioning}, showing sustained exploration compared to PPO. (b) Mean advantage weight applied to the distillation objective across environments, illustrating emphasis on high-quality transitions. (c) Critic agreement fraction between online and frozen reference critics, demonstrating selective but persistent alignment throughout training.}
    \label{fig:finding}
\end{figure}

To understand the mechanisms underlying ROAD-VLA, we analyze internal training dynamics rather than relying solely on final task performance. In particular, we examine how policy entropy, advantage-weighted distillation signals, and critic agreement evolve throughout optimization. These provide insight into why the proposed objective yields robust adaptation under online interaction. \textbf{Policy entropy:} Figure~\ref{fig:finding}(a) illustrates the evolution of policy entropy on \texttt{ER-Repositioning}. While both methods exhibit an initial entropy increase due to PPO's exploration bonus, ROAD-VLA maintains significantly higher entropy at convergence ($3.24$ vs. $3.15$ nats). We interpret this as evidence that advantage-weighted distillation functions as an implicit diversity regularizer. By distilling toward a proximal teacher rather than solely maximizing a point-estimate reward, ROAD-VLA prevents premature policy collapse—a state where the model over-commits to a narrow action mode. Preserving broader action support is critical for OOD robustness; it ensures the policy retains the probabilistic flexibility required to recover from execution errors or perceptual shifts that fall outside the narrow training distribution. \textbf{Advantage weight:} Figure~\ref{fig:finding}(b) reports the mean advantage weight $\bar{\omega}$ applied to the distillation objective. Across both environments, $\bar{\omega}$ shifts from slightly negative values ($-0.033$) toward positive values at convergence ($+0.037$--$+0.040$), exceeding zero on $56\%$ of optimization steps. This trend indicates that as the policy improves, distillation increasingly emphasizes high-quality transitions. This selective emphasis distinguishes the method from uniform distillation, which treats all transitions equally. \textbf{Critic agreement:} Figure~\ref{fig:finding}(c) measures agreement between the online critic and the frozen reference critic on the sign of the advantage estimate. Agreement decreases from approximately $90\%$ early in training to $71$--$75\%$ at convergence, while remaining substantially above chance. This reflects growing policy divergence from the reference checkpoint, alongside continued utility of the reference signal. The agreement gate therefore serves its intended purpose: applying the mixed advantage $\hat A_t^{\mathrm{mix}}$ only under consensus, while reverting to the online advantage otherwise. This mechanism preserves useful reference guidance without introducing stale or conflicting gradients as training progresses.

\begin{wraptable}{r}{0.6\textwidth}
\vspace{-10pt}
\centering
\small
\caption{Ablation study on \texttt{VR-UnseenTable} OOD (mean $\pm$ std).}
\label{tab:ablation}
\begin{tabular}{lc}
\toprule
\textbf{Method}  & \textbf{Success Rate (\%)} \\
\midrule
PPO                          &  87.2 $\pm$ 3.6 \\
\midrule
ROAD-VLA (RelSpatial PI)          &   4.68 $\pm$ 0.0   \\
ROAD-VLA (Plan+RelSpatial PI)          &   4.68 $\pm$ 0.0   \\
ROAD-VLA (MCTS PI)          &   75.8 $\pm$ 2.0   \\
ROAD-VLA (JSD loss)         &   85.9 $\pm$ 1.5 \\
% ROAD-VLA ($\alpha$=0.0)     &   88.5 $\pm$ 2.2  \\
ROAD-VLA (w/o gated)    &  89.8 $\pm$ 0.1  \\
\midrule
\textbf{ROAD-VLA (full)}   & \textbf{91.5 $\pm$ 1.2} \\
\bottomrule
\end{tabular}
\vspace{-10pt}
\end{wraptable}

\subsection{Ablation Study.}
\label{sec:ablation_study}

To evaluate the ROAD-VLA framework, we conducted controlled ablations in the \texttt{VR-UnseenTable} environment across three random seeds (Table~\ref{tab:ablation}). Replacing our advantaged PI with a text-only PI (see Section~\ref{sec:Text_pi}) triggered a performance collapse to 75.8\%—11.4 points below the PPO baseline—confirming that advantage-guided signals are a prerequisite for effective distillation. This failure suggests that post-training VLA weakens the general reasoning of the LLM backbone and that a modality gap prevents the discrete text from providing the precise grounding required for continuous control. Substituting Forward KL with Jensen-Shannon Divergence (JSD loss) reduced success to 85.9\%, as JSD’s compromise between mode-seeking and mode-covering provides a weaker adaptation signal than the mode-covering Forward KL. Finally, removing the sign-agreement gate (w/o gated) resulted in a 1.7-point decrease to 89.8\%, highlighting its role in preventing "distillation inversion" from conflicting critic signals. Varying the reference critic mixing weight $\alpha$ (see Figure~\ref{fig:alpha_ablation}) reveals a stability–adaptability tradeoff: $\alpha{=}1.0$ (frozen reference only) converges rapidly but stagnates as the static critic grows stale, $\alpha{=}0.0$ (online only) adapts continuously yet lacks early-training stability, and $\alpha{=}0.5$ best balances both signals, sustaining reliable gating throughout training and achieving the highest OOD success rate at convergence. Collectively, the full model achieves 91.5 $\pm$ 1.2\%, a 4.3-point gain over the baseline, validating the effectiveness of these complementary components.

% Additionally, removing the reference critic pretrained ($\alpha=0.0$) reduced the success to 88.5\%, indicating that starting from scratch introduces high variance that is otherwise stabilized by our mixing strategy $\alpha=0.5$.

% \begin{table}[t]
% \centering
% \small
% \caption{Ablation study on the \texttt{VR-UnseenTable} environment (OOD success rate,
%          mean $\pm$ std over seeds). Each row isolates one design choice of ROAD-VLA.}
% \label{tab:ablation}
% \begin{tabular}{lccc c}
% \toprule
% \textbf{Method}  & \textbf{ID Success (\%)} \\
% \midrule
% PPO                          &  87.2 $\pm$ 3.6 \\
% \midrule
% ROAD-VLA (Text PI)          &   75.8 $\pm$ 2.0   \\
% ROAD-VLA (JSD loss)         &   85.9 $\pm$ 1.5 \\
% ROAD-VLA ($\alpha$=0.0)     &   88.5 $\pm$ 2.2  \\
% ROAD-VLA ($\alpha$=0.5,w/o gated)    &  89.8 $\pm$ 0.1  \\
% \midrule
% \textbf{ROAD-VLA (full)}   & \textbf{91.5 $\pm$ 1.2} \\
% \bottomrule
% \end{tabular}
% \end{table}

\section{Conclusion}
\label{sec:Conclusion}

In this work, we introduce ROAD-VLA, a novel advantage-guided self-distillation framework designed to mitigate the instability and supervision sparsity inherent in traditional RL for VLA models. By moving beyond intuitive but often ineffective text-based distillation, we demonstrate that converting scalar advantage estimates into dense, token-level supervision provides a significantly more robust adaptation signal. Empirical results on robotic benchmarks validate the superiority of the ROAD-VLA adaptation loop, consistently outperforming the standard PPO baseline across both in-distribution and out-of-distribution environments. Ultimately, this approach offers a scalable path for fine-tuning large-scale embodied agents, enabling them to refine their physical precision while preserving the foundational knowledge of the pretrained backbone. Despite these promising results, several limitations remain. Our experiments focus on pick-and-place manipulation tasks in simulation, and future work should validate ROAD-VLA on physical robots, longer-horizon tasks, and broader task distributions. The framework also depends on a reference PPO critic whose quality can degrade under large distribution shifts, motivating future exploration of critic-free or uncertainty-aware teacher construction.

\section*{Acknowledgements}

This work is supported by the Air Force Office of Scientific Research under award number and ack first to FA2386-24-1-4031. The authors acknowledge the National Computational Infrastructure (NCI Australia), an NCRIS-enabled capability supported by the Australian Government, for providing computational resources used in this study. The authors also acknowledge the Katana computational cluster, supported by Research Technology Services at UNSW Sydney.

{\small
\bibliographystyle{plainnat}
\bibliography{references_cite}
}

%%%%%%%%%%%%%%%%%%%%%%%%%%%%%%%%%%%%%%%%%%%%%%%%%%%%%%%%%%%%
\newpage

\appendix

\section{More Theoretical Results}

\subsection{Derivation of the advantage-guided proximal teacher}
\label{app:prox_teacher_derivation}

This appendix derives the closed-form teacher used in Sec.~\ref{sec:adv_guided_context}
and shows that the advantage-guided logit perturbation is the exact solution of a
KL-regularized reward-shaping problem, applied per action token.

\paragraph{Setup.}
Fix a decoding state $h_t=(o_{\le t},\,l,\,\hat a_{<t})$ and write
$\pi_\theta(\cdot\mid h_t)$ for the current student policy over an action space
$\mathcal A$. The advantage-guided teacher maximizes a shaping reward while staying
proximal to the student:
\begin{equation}
q_t^\star
=
\mathrm{arg\,max}_{q\in\Delta(\mathcal A)}
\Bigl\{
\underbrace{\mathbb E_{a\sim q}\bigl[r_t(a)\bigr]}_{\mathrm{local~improvement}}
-\;
\tau\,\underbrace{\mathrm{KL}\!\bigl(q\,\|\,\pi_\theta(\cdot\mid h_t)\bigr)}_{\mathrm{stay~near~}\pi_\theta}
\Bigr\},
\end{equation}
where $r_t\colon\mathcal A\to\mathbb R$ is an advantage-derived shaping reward, $\tau>0$ is
a temperature controlling proximity to $\pi_\theta$, and $\Delta(\mathcal A)$ is the
probability simplex. We first solve this problem for a generic finite $\mathcal A$, then
instantiate it at the token level.

\paragraph{Closed-form solution.}
The objective is strictly concave in $q$ (relative entropy is strictly convex in its first
argument and $\tau>0$), so it admits a unique maximizer over the simplex, given by the
exponential tilt of the student:
\begin{equation}
q_t^\star(a)
=
\frac{\pi_\theta(a\mid h_t)\,\exp\!\bigl(r_t(a)/\tau\bigr)}
{\sum_{a'\in\mathcal A}\pi_\theta(a'\mid h_t)\,\exp\!\bigl(r_t(a')/\tau\bigr)}
\;\propto\;
\pi_\theta(a\mid h_t)\,\exp\!\bigl(r_t(a)/\tau\bigr).
\end{equation}
Writing the student in logit form, $\pi_\theta(\cdot\mid h_t)=\mathrm{softmax}(z_t)$, the
solution is a simple shift of the logits,
\begin{equation}
q_t^\star=\mathrm{softmax}\!\Bigl(z_t+\frac{1}{\tau}\,r_t\Bigr),
\qquad r_t\in\mathbb R^{|\mathcal A|},
\end{equation}
so shaping the reward by $r_t$ is equivalent to perturbing the logits by $r_t/\tau$.

\paragraph{Proof.}
Expanding the relative entropy, the objective reads
$\sum_{a} q(a)\,r_t(a)-\tau\sum_{a} q(a)\log\frac{q(a)}{\pi_\theta(a\mid h_t)}$.
With a multiplier $\lambda$ for $\sum_a q(a)=1$, the Lagrangian is
\begin{equation}
\mathcal J(q,\lambda)
=\sum_{a} q(a)\,r_t(a)
-\tau\sum_{a} q(a)\log\frac{q(a)}{\pi_\theta(a\mid h_t)}
+\lambda\Bigl(\sum_{a} q(a)-1\Bigr).
\end{equation}
Stationarity in $q(a)$ gives
\begin{equation}
r_t(a)-\tau\Bigl(\log\frac{q(a)}{\pi_\theta(a\mid h_t)}+1\Bigr)+\lambda=0
\;\Longrightarrow\;
\log\frac{q(a)}{\pi_\theta(a\mid h_t)}=\frac{r_t(a)}{\tau}+\frac{\lambda-\tau}{\tau}.
\end{equation}
The last term is constant in $a$, so $q(a)=C\,\pi_\theta(a\mid h_t)\exp(r_t(a)/\tau)$;
enforcing $\sum_a q(a)=1$ fixes $C^{-1}=\sum_{a'}\pi_\theta(a'\mid h_t)\exp(r_t(a')/\tau)$,
which is the claimed solution. The entropy term forces full support
($q_t^\star(a)>0$ wherever $\pi_\theta>0$), so the non-negativity constraints are inactive
and the stationary point is the global maximizer. 

\paragraph{Per-token instantiation.}
During on-policy adaptation we obtain a single advantage estimate for the sampled action
$\hat a_t$, summarized by the scalar weight $\omega_t\in\mathbb R$ of
Sec.~\ref{sec:adv_guided_context}. Rather than solving one proximal problem over the full
joint action space $\mathcal A=\prod_{k=1}^K\mathcal V_k$, we apply the construction above
\emph{independently at each of the $K=7$ OpenVLA action tokens}. At position $k$, condition
on the sampled prefix $\hat a_{<k}$ and write the student token conditional as
$p^\theta_{t,k}(\cdot\mid\hat a_{<k},h_t)=\mathrm{softmax}(z_{t,k})$. With the one-point
token reward
\begin{equation}
r_{t,k}(u)
=
\tau\eta\,\omega_t\,\mathbf 1[u=\hat a_{t,k}],
\qquad u\in\mathcal V_k,
\end{equation}
the closed-form solution specializes to
\begin{equation}
q_{t,k}^\star
=
\mathrm{softmax}\!\bigl(z_{t,k}+\eta\,\omega_t\,e_{\hat a_{t,k}}\bigr),
\qquad
q_{t,k}^\star(u)
=
\frac{\exp\!\bigl(z_{t,k}(u)+\eta\,\omega_t\,\mathbf 1[u=\hat a_{t,k}]\bigr)}
{\sum_{v\in\mathcal V_k}\exp\!\bigl(z_{t,k}(v)+\eta\,\omega_t\,\mathbf 1[v=\hat a_{t,k}]\bigr)},
\end{equation}
where $e_{\hat a_{t,k}}$ is the one-hot vector of the sampled token and the prefactor
$\tau\eta$ makes the tilt $\exp(r_{t,k}/\tau)$ a logit shift of size $\eta\omega_t$. We
use $\tau=1$ and $\eta=1/\tau=1$ throughout, matching Sec.~\ref{sec:adv_guided_context}.
Crucially, $\omega_t$ is \emph{signed}: a positive weight shifts mass toward the sampled
token, a negative weight shifts it away, and the magnitude sets how far the teacher departs
from the student. No ReLU or positive clipping is applied; the ablation in
Appendix~\ref{subsec:results_adv_sign_pos_gate} shows the signed weight performs on par with
the non-negative variant.

\paragraph{Autoregressive teacher.}
The token teachers compose into a single autoregressive distribution,
\begin{equation}
q_t^\star(a\mid h_t)
=
\prod_{k=1}^{K}
q_{t,k}^\star(a_k\mid a_{<k},h_t),
\end{equation}
which is a valid policy over $\mathcal A$ and is the teacher distilled in the token-level
objective $\mathcal L_{\mathrm{AGD}}$. Because the teacher is defined token-wise against the
student conditionals $p^\theta_{t,k}$, the teacher-to-student relative entropy decomposes
exactly by the chain rule,
\begin{equation}
\mathrm{KL}\!\bigl(q_t^\star\,\|\,\pi_\theta\bigr)
=
\sum_{k=1}^{K}
\mathbb E_{a_{<k}\sim q_t^\star}
\Bigl[
\mathrm{KL}\!\bigl(q_{t,k}^\star(\cdot\mid a_{<k},h_t)\,\|\,p^\theta_{t,k}(\cdot\mid a_{<k},h_t)\bigr)
\Bigr],
\end{equation}
which is the decomposition invoked in Theorem~1. The construction thus expands a scalar
rollout advantage into dense token-wise targets while keeping each token teacher proximal to
the corresponding student conditional.

\paragraph{Per-token vs.\ joint proximal.}
Applying the proximal solution per token does not in general coincide with solving the joint
problem over $\mathcal A$ and then conditioning: the joint solution normalizes over all token
combinations with a single partition function, whereas the per-token teacher normalizes each
position separately. We adopt the per-token construction by design---it is the object that
$\mathcal L_{\mathrm{AGD}}$ optimizes and the one for which the chain-rule decomposition above
holds exactly, requiring no independence assumption beyond conditioning on the sampled prefix.

\subsection{Proof of Theorem~1}
\label{app:proof_self_privileged_adv_distillation}

We use two facts from Appendix~\ref{app:prox_teacher_derivation}: at each position $k$, the
teacher $q_{t,k}^\star(\cdot\mid a_{<k},h_t)$ is the KL-proximal solution obtained by
exponentially tilting the student conditional $p^\theta_{t,k}(\cdot\mid a_{<k},h_t)$ with the
token reward $r_{t,k}$; and the autoregressive teacher-policy KL decomposes by the chain rule
into a sum of token-level KL terms.

Fix a timestep $t$ and a context $h_t$. Since the token reward $r_{t,k}$ may be signed,
center it under the student conditional,
\begin{equation}
\bar r_{t,k}(u\mid a_{<k},h_t)
=
r_{t,k}(u)
-
\mathbb E_{v\sim p^\theta_{t,k}(\cdot\mid a_{<k},h_t)}\!\bigl[r_{t,k}(v)\bigr].
\end{equation}
For each prefix $a_{<k}$ the subtracted term is independent of the token value $u$, so it
cancels under the softmax and leaves the token teacher unchanged; moreover
$\mathbb E_{u\sim p^\theta_{t,k}}[\bar r_{t,k}(u\mid a_{<k},h_t)]=0$.

\emph{Per-token improvement.}
The student conditional $p^\theta_{t,k}$ is feasible for the per-token proximal problem, with
objective value $\mathbb E_{p^\theta_{t,k}}[\bar r_{t,k}]-\tau\cdot 0=0$. Since $q_{t,k}^\star$
is the maximizer, its objective value is at least as large,
\begin{equation}
\mathbb E_{u\sim q_{t,k}^\star}\!\bigl[\bar r_{t,k}(u\mid a_{<k},h_t)\bigr]
-
\tau\,\mathrm{KL}\!\bigl(q_{t,k}^\star(\cdot\mid a_{<k},h_t)\,\|\,p^\theta_{t,k}(\cdot\mid a_{<k},h_t)\bigr)
\;\ge\;0,
\end{equation}
hence
\begin{equation}
\mathbb E_{u\sim q_{t,k}^\star}\!\bigl[\bar r_{t,k}\bigr]
\;\ge\;
\tau\,\mathrm{KL}\!\bigl(q_{t,k}^\star(\cdot\mid a_{<k},h_t)\,\|\,p^\theta_{t,k}(\cdot\mid a_{<k},h_t)\bigr).
\end{equation}

\emph{Aggregation.}
Define the centered sequence reward
$\bar r_t(a)=\sum_{k=1}^{K}\bar r_{t,k}(a_k\mid a_{<k},h_t)$, which is exactly the centered
reward $\bar r_t$ of Theorem~1 evaluated token by token (each token reward $r_{t,k}$ is
evaluated at the action's $k$-th component $a_k$). Taking expectation over prefixes
$a_{<k}\sim q_t^\star$, summing the per-token inequality over $k$, and applying the
chain-rule decomposition of the teacher-policy KL,
\begin{equation}
\mathbb E_{a\sim q_t^\star(\cdot\mid h_t)}\!\bigl[\bar r_t(a)\bigr]
\;\ge\;
\tau\,\mathrm{KL}\!\bigl(q_t^\star(\cdot\mid h_t)\,\|\,\pi_\theta(\cdot\mid h_t)\bigr).
\end{equation}

\emph{Calibration.}
By the advantage-alignment assumption under the teacher,
\begin{equation}
\mathbb E_{a\sim q_t^\star}\!\bigl[A_t^{\pi_\theta}(h_t,a)\bigr]
\;\ge\;
\beta\,\mathbb E_{a\sim q_t^\star}\!\bigl[\bar r_t(a)\bigr]-\epsilon_{\mathrm{cal}}
\;\ge\;
\beta\tau\,\mathrm{KL}\!\bigl(q_t^\star(\cdot\mid h_t)\,\|\,\pi_\theta(\cdot\mid h_t)\bigr)
-\epsilon_{\mathrm{cal}}.
\end{equation}

\emph{Transfer to the student.}
Write $D_t^{\mathrm{dist}}(h_t)=\mathrm{KL}(q_t^\star(\cdot\mid h_t)\,\|\,\pi_{\theta'}(\cdot\mid h_t))$.
By Pinsker's inequality, the total variation between teacher and student is bounded as
\begin{equation}
\mathrm{TV}\!\bigl(q_t^\star,\pi_{\theta'}\bigr)
\;\le\;
\sqrt{\frac12\,D_t^{\mathrm{dist}}(h_t)}.
\end{equation}
Since $|A_t^{\pi_\theta}(h_t,a)|\le B$, the advantage expectation differs between the two
distributions by at most
\begin{equation}
\bigl|
\mathbb E_{a\sim\pi_{\theta'}}\!\bigl[A_t^{\pi_\theta}(h_t,a)\bigr]
-
\mathbb E_{a\sim q_t^\star}\!\bigl[A_t^{\pi_\theta}(h_t,a)\bigr]
\bigr|
\;\le\;
2B\,\mathrm{TV}\!\bigl(q_t^\star,\pi_{\theta'}\bigr)
\;\le\;
\sqrt2\,B\,\sqrt{D_t^{\mathrm{dist}}(h_t)}.
\end{equation}
With $C=\sqrt2$, combining this with the calibration bound yields
\begin{equation}
\mathbb E_{a\sim\pi_{\theta'}}\!\bigl[A_t^{\pi_\theta}(h_t,a)\bigr]
\;\ge\;
\beta\tau\,\mathrm{KL}\!\bigl(q_t^\star(\cdot\mid h_t)\,\|\,\pi_\theta(\cdot\mid h_t)\bigr)
-\epsilon_{\mathrm{cal}}
-CB\sqrt{D_t^{\mathrm{dist}}(h_t)}.
\end{equation}

\emph{Performance difference.}
The finite-horizon performance-difference lemma gives
\begin{equation}
J(\pi_{\theta'})-J(\pi_\theta)
=
\frac{1}{T}\sum_{t=1}^{T}
\mathbb E_{h_t\sim d_t^{\pi_{\theta'}}}\,
\mathbb E_{a\sim\pi_{\theta'}(\cdot\mid h_t)}\!\bigl[A_t^{\pi_\theta}(h_t,a)\bigr].
\end{equation}
Substituting the per-state lower bound yields
\begin{equation}
J(\pi_{\theta'})
\;\ge\;
J(\pi_\theta)
+
\frac{1}{T}\sum_{t=1}^{T}
\mathbb E_{h_t\sim d_t^{\pi_{\theta'}}}
\Bigl[
\beta\tau\,\mathrm{KL}\!\bigl(q_t^\star(\cdot\mid h_t)\,\|\,\pi_\theta(\cdot\mid h_t)\bigr)
-CB\sqrt{D_t^{\mathrm{dist}}(h_t)}
\Bigr]
-\epsilon_{\mathrm{cal}},
\end{equation}
which is the claimed bound. 

\section{ROAD-VLA Pseudocode}
\begin{algorithm}[h]
\caption{\textsc{ROAD-VLA}: Advantage-Guided Online Self-Distillation}
\label{alg:roadvla}
\begin{algorithmic}[1]
\Require Student policy $\pi_\theta$, pretrained PPO critic $V^{\mathrm{ref}}$,
         perturbation strength $\eta=1$, mixing coefficient $\alpha=0.5$,
         clipping bound $c=2$, learning rate $\gamma$, language instruction $l$
\Ensure Updated student policy $\pi_{\theta'}$
\For{each training iteration}
    \State \textcolor{gray}{\textit{// Step 1: Collect on-policy rollout}}
    \State Sample trajectory $\hat{\tau} = \{(o_1, \hat{a}_1), \dots, (o_T, \hat{a}_T)\} \sim \pi_\theta(\cdot \mid l)$
    \State Observe episode reward $R(\hat{\tau})$
    \For{each timestep $t = 1, \dots, T$}
        \State \textcolor{gray}{\textit{// Step 2: Compute calibrated, agreement-gated advantage}}
        \State Compute intrinsic advantage $\hat{A}_t^{\mathrm{int}}$ from current critic
        \State Compute reference advantage $\hat{A}_t^{\mathrm{ref}}$ from $V^{\mathrm{ref}}$
        \State Calibrate: $\tilde{A}_t^{\mathrm{ref}} = \mu_{\mathrm{int}} + \frac{\sigma_{\mathrm{int}}}{\sigma_{\mathrm{ref}} + \varepsilon}\bigl(\hat{A}_t^{\mathrm{ref}} - \mu_{\mathrm{ref}}\bigr)$
        \If{$\mathrm{sign}(\hat{A}_t^{\mathrm{int}}) = \mathrm{sign}(\tilde{A}_t^{\mathrm{ref}})$} \textcolor{gray}{\textit{// agreement gate (on calibrated ref)}}
            \State $\hat{A}_t^{\mathrm{mix}} = \hat{A}_t^{\mathrm{int}} + \alpha\bigl(\tilde{A}_t^{\mathrm{ref}} - \hat{A}_t^{\mathrm{int}}\bigr)$
        \Else
            \State $\hat{A}_t^{\mathrm{mix}} = \hat{A}_t^{\mathrm{int}}$
        \EndIf
        \State Perturbation weight (signed, no ReLU):
               $\omega_t = \mathrm{clip}\!\left(\frac{\hat{A}_t^{\mathrm{mix}} - \mu_{\mathrm{mix}}}{\sigma_{\mathrm{mix}} + \varepsilon},\, -c,\, c\right)$
        \For{each action token $k = 1, \dots, K$}
            \State \textcolor{gray}{\textit{// Step 3: Construct advantage-guided teacher}}
            \State Retrieve student logits $z_{t,k}$ and sampled token $\hat{a}_{t,k}$
            \State Perturb logits: $q^\star_{t,k} = \mathrm{softmax}\bigl(z_{t,k} + \eta\,\omega_t\, e_{\hat{a}_{t,k}}\bigr)$
        \EndFor
    \EndFor
    \State \textcolor{gray}{\textit{// Step 4: Distil student toward the local teacher}}
    \State Compute loss:
    $\mathcal{L}_{\mathrm{AGD}} = \frac{1}{TK}\sum_{t=1}^{T}\sum_{k=1}^{K} \mathrm{KL}\bigl(q^\star_{t,k} \,\|\, p^\theta_{t,k}\bigr)$
    \State Update $\theta \leftarrow \theta - \gamma\,\nabla_\theta \mathcal{L}_{\mathrm{AGD}}$
\EndFor
\end{algorithmic}
\end{algorithm}

\section{More Experimental Settings}
\subsection{Environment Details}
\label{app:env_details}

In this section, we provide the technical specifications for the evaluation suite used in Section~\ref{sec:Experiments}. All environments are obtained from the RL4VLA repo~\footnote{\url{https://github.com/gen-robot/RL4VLA}}, which provides a standardized interface for reinforcement learning with vision-language-action models. Observations consist of RGB images from a front-view camera and the robot's proprioceptive state, while the action space is discretized into $K=7$ tokens per step following the OpenVLA specification.

\begin{table}[h]
\centering
\caption{Mapping of main paper environment aliases to technical identifiers.}
\label{tab:env_mapping}
\small
\begin{tabular}{l l l}
\toprule
\textbf{Category} & \textbf{Main Paper Alias} & \textbf{Simulator Environment ID} \\
\midrule
\multirow{3}{*}{Visual Robustness} & \texttt{VR-UnseenTable} & \texttt{PutOnPlateInScene25VisionImage-v1} \\
 & \texttt{VR-DynamicTexture} & \texttt{PutOnPlateInScene25VisionTexture05-v1} \\
 & \texttt{VR-DynamicNoise} & \texttt{PutOnPlateInScene25VisionWhole05-v1} \\
\midrule
\multirow{3}{*}{Compositional Reasoning} & \texttt{CR-MultiObject} & \texttt{PutOnPlateInScene25MultiCarrot-v1} \\
 & \texttt{CR-MultiReceptacle} & \texttt{PutOnPlateInScene25MultiPlate-v1} \\
\midrule
\multirow{2}{*}{Execution Robustness} & \texttt{ER-InitPose} & \texttt{PutOnPlateInScene25EEPose-v1} \\
 & \texttt{ER-Repositioning} & \texttt{PutOnPlateInScene25PositionChangeTo-v1} \\
\bottomrule
\end{tabular}
\end{table}

\subsubsection{Perturbation Protocols}

To systematically probe the robustness of ROAD-VLA, we implemented the following perturbations across our axes of evaluation:

\paragraph{Visual Robustness.} 
These environments test the model's invariance to low-level visual shifts.
\begin{itemize}
    \item \textbf{Unseen Table:} Replaces the training table surface with diverse, unseen textures and materials (e.g., wood, marble, metallic) not present in the warm-up data.
    \item \textbf{Dynamic Texture:}  A random distractor texture is overlaid on the object, receptacle, and robot arm. This texture is re-cropped and resized at \textbf{every timestep} to create non-stationary visual noise. We evaluate at the strong intensity levels: \textit{strong} ($\alpha=0.5$).
    \item \textbf{Dynamic Noise:}  Similar to dynamic textures, but the flickering noise is overlaid on the entire image frame, simulating extreme sensor interference or low-light artifacts at intensities of $0.5$.
\end{itemize}

\paragraph{Compositional Reasoning.} 
These environments test semantic grounding and the ability to handle distractors.
\begin{itemize}
    \item \textbf{Multi-Object/Receptacle:} Spawns multiple distractor objects (e.g., additional carrots or plates) that were not present during the zero-shot task phase. The model must correctly identify the target specified in the instruction $l$ despite visual similarity.
\end{itemize}

\paragraph{Execution Robustness.} 
These environments test temporal adaptation and dynamic error correction.
\begin{itemize}
    \item \textbf{Init Pose:} Randomizes the initial end-effector (EE) pose outside the standard training distribution, requiring the model to generalize its reach-to-grasp trajectories.
    \item \textbf{Repositioning (Mid-episode):} To test for closed-loop visual feedback, the target object is teleported to a new random position at the 5th timestep of the episode. This requires the model to break its initial plan and adapt to the updated state mid-execution.
\end{itemize}

\subsection{Implementation Details}
\label{app:imp_details}

\subsubsection{Model Initialization and Warm-up}
To bypass the high variance of learning from scratch in robotic manipulation, all policies (both ROAD-VLA and PPO) are initialized from a shared warm-up checkpoint.

Source: The checkpoint is sourced from the website~\footnote{\url{https://huggingface.co/collections/gen-robot/rlvla}}.

Data: This model was fine-tuned on a small-scale dataset of 140 expert trajectories, which were generated using a motion planner (e.g., octo-small) to demonstrate basic prehension and reaching primitives.

Parameters: We utilize LoRA (Low-Rank Adaptation) for all reinforcement learning experiments to ensure memory efficiency. The adapters are applied to the linear layers of the vision-language backbone.

\subsubsection{Optimization, Training, and Hyperparameters}
We implement our training pipeline using the \texttt{AdamW} optimizer with a policy learning rate of $1 \times 10^{-4}$ and momentum parameters $\beta_1=0.9, \beta_2=0.999$. To stabilize the high-dimensional gradient updates inherent in VLA adaptation, we apply a global gradient clipping threshold of $1.0$ and utilize a cosine learning rate schedule. 

\textbf{Parallelized Rollouts.} We utilize a highly parallelized sampling strategy with $64$ concurrent environments, collecting rollouts with a fixed episode horizon of $80$ steps. To maximize throughput on NVIDIA H100/H200 GPUs, we employ a rollout inference batch size of $32$. The total environment interaction budget is capped at $850,000$ steps per experiment, which provides sufficient coverage for both in-distribution and out-of-distribution adaptation.

\textbf{PPO Configuration.} We utilize PPO with Generalized Advantage Estimation (GAE) for variance reduction ($\gamma=0.99, \lambda=0.95$). Each update consists of a single PPO epoch per rollout to prevent aggressive policy drift. To manage the memory footprint of the $7$B-parameter model, we use a minibatch size of $8$ combined with $20$ gradient accumulation steps, resulting in an effective batch size that balances stability and compute efficiency.

\textbf{ROAD-VLA Specifics.} In ROAD-VLA, the advantage-guided distillation loss is applied with a coefficient of $0.5$ using either JSD or a forward KL-divergence objective. The teacher distribution is constructed by applying a logit bump scale of $\eta=1.0$ to actions within a clipped advantage range of $2.0$. We integrate the Reference Agreement Gate to modulate the influence of the frozen PPO critic, maintaining a mixing weight of $\alpha=0.5$ between the online and reference value signals. We also have experimented with $\alpha=0.0$ and $\alpha=1.0$. During training, we maintain a sampling temperature of $1.0$ to encourage exploration, which is reduced to $0.6$ during evaluation for deterministic performance.

See summary of key hyperparameters in Table.~\ref{tab:hyperparams}.

\begin{table}[h]
\centering
\caption{Hyperparameters used for all experiments.}
\label{tab:hyperparams}
\begin{tabular}{lc}
\toprule
\textbf{Hyperparameter} & \textbf{Value} \\
\midrule
\multicolumn{2}{l}{\textit{Shared (PPO and ROAD-VLA)}} \\
\midrule
Parallel environments          & 64 \\
Episode length (steps)         & 80 \\
Total environment steps        & 850{,}000 \\
Rollout inference batch size   & 32 \\
PPO minibatch size             & 8 \\
Gradient accumulation steps    & 20 \\
PPO epochs per rollout         & 1 \\
GAE discount factor ($\gamma$) & 0.99 \\
GAE $\lambda$                  & 0.95 \\
Policy LoRA rank               & 32 \\
Policy learning rate           & $1 \times 10^{-4}$ \\
Adam $\beta_1$ / $\beta_2$     & 0.9 / 0.999 \\
Entropy coefficient            & 0.01 \\
Sampling temperature (train)   & 1.0 \\
Sampling temperature (eval)    & 0.6 \\
\midrule
\multicolumn{2}{l}{\textit{ROAD-VLA only}} \\
\midrule
Distillation loss coefficient  & 0.5 \\
Logit bump scale ($\eta$)      & 1.0 \\
Advantage clip range ($c$)         & 2.0 \\
Distillation temperature       & 1.0 \\
Divergence type                & Forward KL or JSD \\
Global token weight            & 1.0 \\
Positive-only advantage gate   & False \\
Reference agreement gate       & True \\
Reference critic type          & PPO value head \\
Reference critic mixing weight ($\alpha$) & 0.0,0.5,1.0 \\
\bottomrule
\end{tabular}
\end{table}

\section{More Analyses and Ablation Studies}

\subsection{Failure Mode Analysis}
\label{sec:failure_mode_analysis}

To better understand model behavior, we examine the dominant failure patterns under representative OOD settings and compare how ROAD-VLA alters these outcomes. 

In execution robustness tasks, in particular \texttt{ER-Repositioning}, PPO commonly exhibits a grasp-then-drift behavior: after a successful grasp, the policy moves toward the original training-distribution plate location rather than the shifted target, leading to object drop or task failure. As illustrated in Figure~\ref{fig:qualitative_evident}, this reflects reliance on a fixed spatial prior rather than continuous visual correction. ROAD-VLA reduces this failure pattern by maintaining stronger alignment with updated task cues, enabling successful object placement under shifted configurations. 

In visual robustness tasks, specifically \texttt{VR-DynamicTexture}, the dominant error is perception-induced no-grasp. PPO frequently fails to localize the object under unseen textures, resulting in repeated unsuccessful approach attempts and eventual timeout. By contrast, ROAD-VLA demonstrates more reliable visual grounding, achieving successful grasps and task completion despite altered appearances. As shown in Figure~\ref{fig:grasp_rate_ood}, ROAD-VLA achieves a higher OOD grasp rate throughout training, peaking at $81.8\%$ at step 149 compared to $81.2\%$ for PPO at step 164 — both a higher peak and faster convergence. At the final checkpoint, ROAD-VLA maintains a $2.0\%$ advantage ($76.0\%$ vs.\ $74.0\%$), with the gap reaching up to $7.3\%$ during mid-training. This suggests improved robustness to appearance-level perturbations, though perceptual invariance remains a challenge in highly shifted scenes. 

Overall, the qualitative comparisons reveal that ROAD-VLA not only improves final success rates, but also changes the underlying failure dynamics from rigid prior-driven execution and unstable perception toward more adaptive, task-aligned behavior.

\subsection{$\alpha$ analysis}
\label{sec:alpha_analysis}
The ablation over the reference critic mixing weight $\alpha$ reveals a consistent stability--adaptability tradeoff across both \texttt{VR-UnseenTable} and \texttt{ER-Repositioning} (see Figure~\ref{fig:alpha_ablation}). Note that all curves are smoothed with rolling average for visual clarity; the values discussed below therefore differ slightly from the exact figures reported in Table~\ref{tab:roadvla_results}. 

Using only the frozen reference critic ($\alpha=1.0$) yields the fastest initial gains: in \texttt{VR-UnseenTable} it starts above both alternatives at roughly 63\% OOD success, and in \texttt{ER-Repositioning} it similarly leads early. This reflects the frozen PPO value head's strong, stable gating signal during the early phase of adaptation. However, the reference signal becomes increasingly stale as the online policy diverges from the frozen critic's training distribution, causing the $\alpha=1.0$ curve to plateau and be overtaken in both environments. 

Relying solely on the online critic ($\alpha=0.0$) produces the opposite pattern: a slower, noisier start as the online critic is initially poorly calibrated, but steadier improvement as it co-adapts with the evolving policy, reaching approximately 78\% in \texttt{VR-UnseenTable} and 67\% in \texttt{ER-Repositioning}. 

The mixed setting ($\alpha=0.5$) achieves the best final performance in both environments despite starting comparably to $\alpha=0.0$. The frozen critic stabilizes gating decisions early while the online critic remains unreliable; as training proceeds, the adaptive online signal compensates for the staleness of the frozen reference. 

This combination yields the highest final OOD success rates of approximately 84\% on \texttt{VR-UnseenTable} and 75\% on \texttt{ER-Repositioning}, supporting $\alpha=0.5$ as the default configuration for balancing early training stability with long-term adaptability.

\subsection{Sensitivity to Reference Critic Checkpoint}
In all main experiments, ROAD-VLA freezes the final PPO checkpoint (episode 165) as the reference critic. To assess the sensitivity of this design choice, we conduct an ablation study in Figure~\ref{fig:PPO warm-up checkpoint}, comparing reference critics frozen at different training stages (episodes 39, 79, 119, and 159) against the default on \texttt{VR-UnseenTable} and \texttt{ER-Repositioning}. Overall, ROAD-VLA is relatively robust to the choice of reference checkpoint, with all five variants following similar learning trajectories and converging within a comparable performance band. From a practical standpoint, using the final checkpoint as the default remains a safe and well-motivated choice: it performs best on visual robustness tasks and competitively on execution robustness tasks, without requiring any additional checkpoint selection. When intermediate checkpoints are available and the target task involves strong spatial or execution shifts, selecting a checkpoint from approximately 50\% of total training steps (episode 79 in our setting) can provide a modest but consistent improvement.

\subsection{Advantage Sign Gate}
\label{subsec:results_adv_sign_pos_gate}

\begin{table}[h]
\centering
\caption{Ablation on the advantage sign gate (\texttt{positive\_only}) on \texttt{CR-MultiObject}.
All other hyperparameters are fixed. Results are mean $\pm$ std over three seeds.}
\label{tab:ablation_relu}
\begin{tabular}{lcc}
\toprule
\textbf{Method} & \textbf{ID Success (\%)} & \textbf{OOD Success (\%)} \\
\midrule
ROAD-VLA (Positive-only advantage gate=TRUE) & 67 $\pm$ 8 & 57 $\pm$ 7 \\
ROAD-VLA (Positive-only advantage gate=FALSE) & 80 $\pm$ 6 & 63 $\pm$ 4 \\
\bottomrule
\end{tabular}
\end{table}

The advantage sign gate controls whether the pseudo-teacher distillation signal is applied bidirectionally --- boosting good actions and suppressing bad ones --- or restricted to positive-advantage actions only. Table~\ref{tab:ablation_relu} reports results on \texttt{CR-MultiObject}. Disabling the gate yields substantially stronger performance on both ID and OOD evaluations, achieving $80 \pm 6$ and $63 \pm 4$ respectively, compared to $67 \pm 8$ and $57 \pm 7$ with the gate enabled. Beyond the raw improvement, the bidirectional setting also exhibits notably lower variance, suggesting more stable training dynamics. This is consistent with the intuition that in tasks, where many action sequences are suboptimal, the negative-advantage signal carries rich information: explicitly suppressing poor actions in the teacher distribution provides strong corrective gradients that the one-sided gate discards entirely. Restricting distillation to positive-advantage actions alone leaves the policy without direct feedback on what not to do. Accordingly, ROAD-VLA adopts Positive-only advantage gate=FALSE as its default configuration.

\subsection{Computational Cost}
\label{app:compute}
We report the wall-clock training time for \textsc{ROAD-VLA} and PPO on the \texttt{VR-DynamicNoise} environment (\texttt{PutOnPlateInScene25VisionWhole05-v1}) as a representative benchmark. Both methods are trained for 850,000 environment steps with 64 parallel environments, a rollout episode length of 80, and identical gradient accumulation settings (20 steps, minibatch size 8).
PPO completes training in approximately \textbf{25.4 hours} on a single NVIDIA H100 80GB GPU. ROAD-VLA completes training in approximately \textbf{40.8 hours} on a single NVIDIA H100 GPU — a $1.6\times$ overhead relative to PPO. This additional cost arises from two sources: (1) the forward pass through the teacher distribution to compute logit perturbations at each on-policy timestep, and (2) the token-level KL distillation loss computed across all K=7 action tokens per step.
Importantly, \textsc{ROAD-VLA} introduces \emph{no additional rollout overhead} — the teacher is constructed analytically from the student's own logits and advantage estimates rather than through a separate forward pass of a distinct model. The extra computation is therefore confined to the gradient update phase.

\section{More Visualizations}
\subsection{Failure Mode Visualization} 
\begin{figure}[h]
    \centering
    \includegraphics[width=0.92\linewidth]{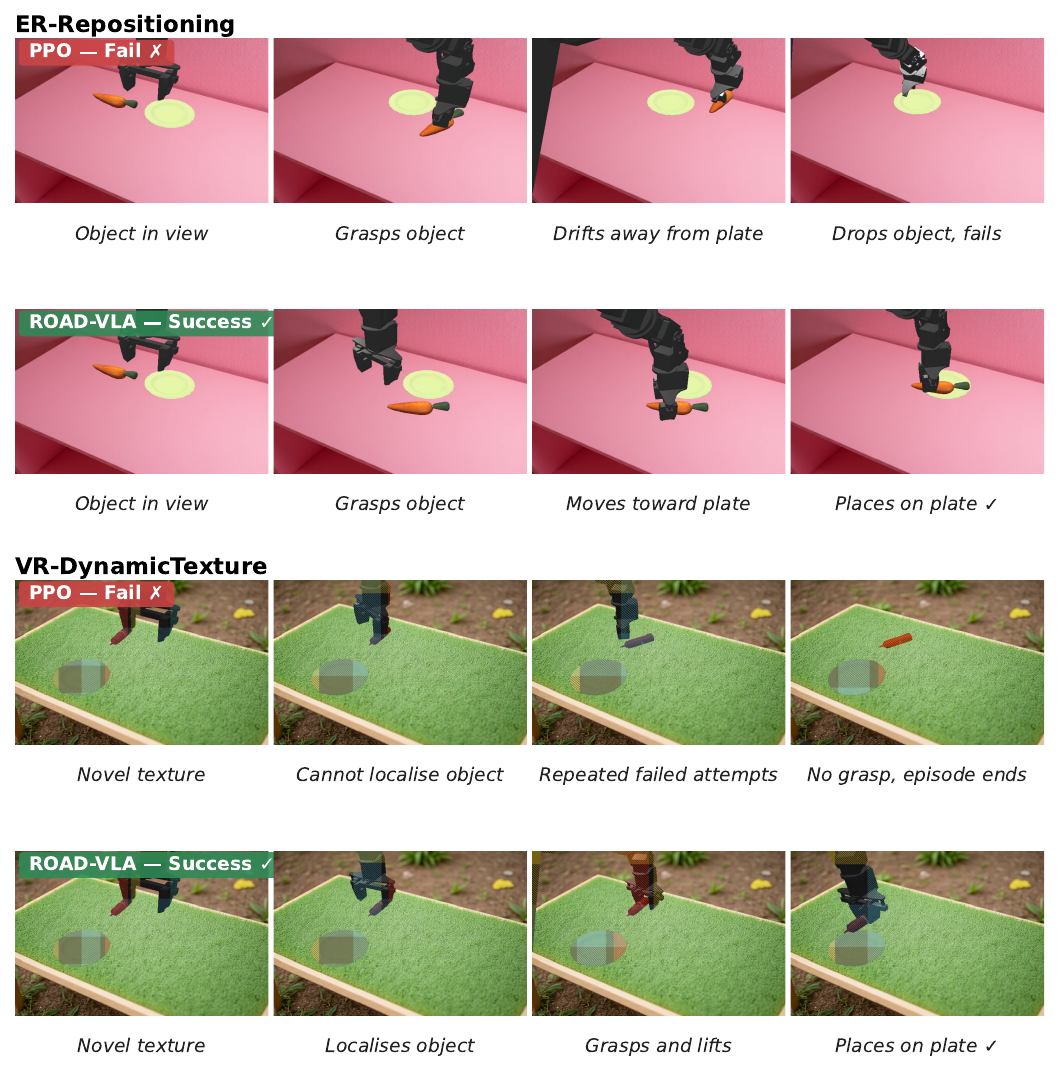}
    \caption{Qualitative rollout comparison under OOD conditions. In \texttt{ER-Repositioning}, PPO drifts toward the training-distribution receptacle location after grasping, while ROAD-VLA successfully adapts and places the object at the shifted target. In \texttt{VR-DynamicTexture}, PPO fails to reliably localize the object under unseen textures, whereas ROAD-VLA maintains stable perception and completes the task.}
    \label{fig:qualitative_evident}
\end{figure}

Refer to Figure.~\ref{fig:qualitative_evident}.

\subsection{Grasp Rate on \texttt{VR} Task} 
\begin{figure}[h]
    \centering
    \includegraphics[width=1\linewidth]{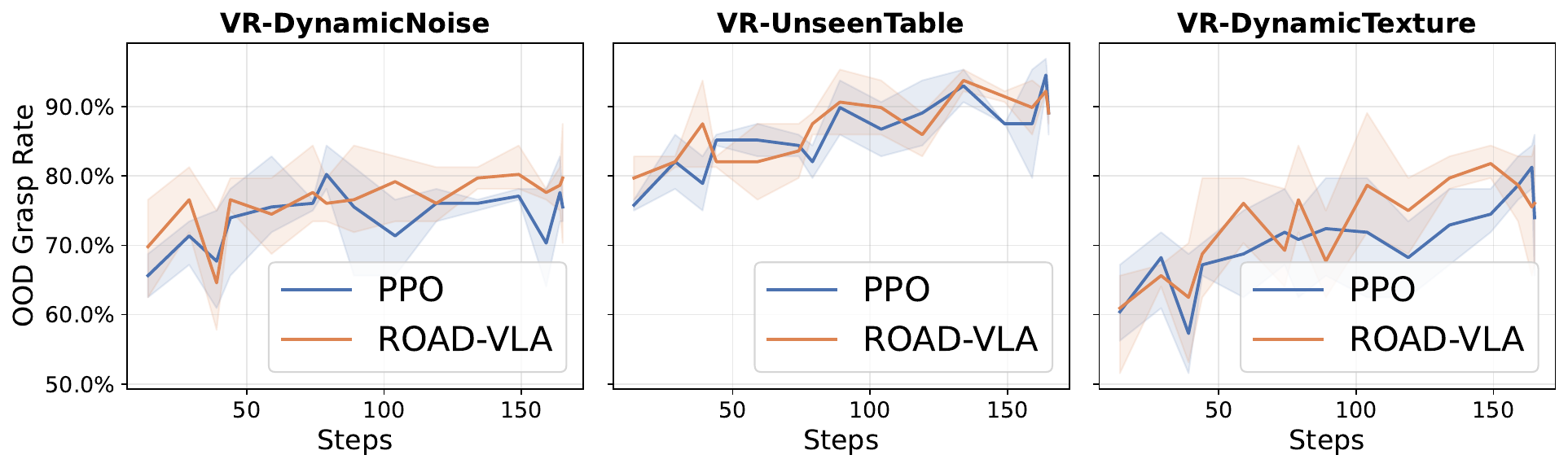}
    \caption{
    OOD grasp success rate on three \texttt{VR} environments. ROAD-VLA reaches its peak grasp rate earlier than PPO and maintains a consistent advantage throughout training.
    }
    \label{fig:grasp_rate_ood}
\end{figure}

\begin{figure}[h]
    \centering
    \includegraphics[width=1\linewidth]{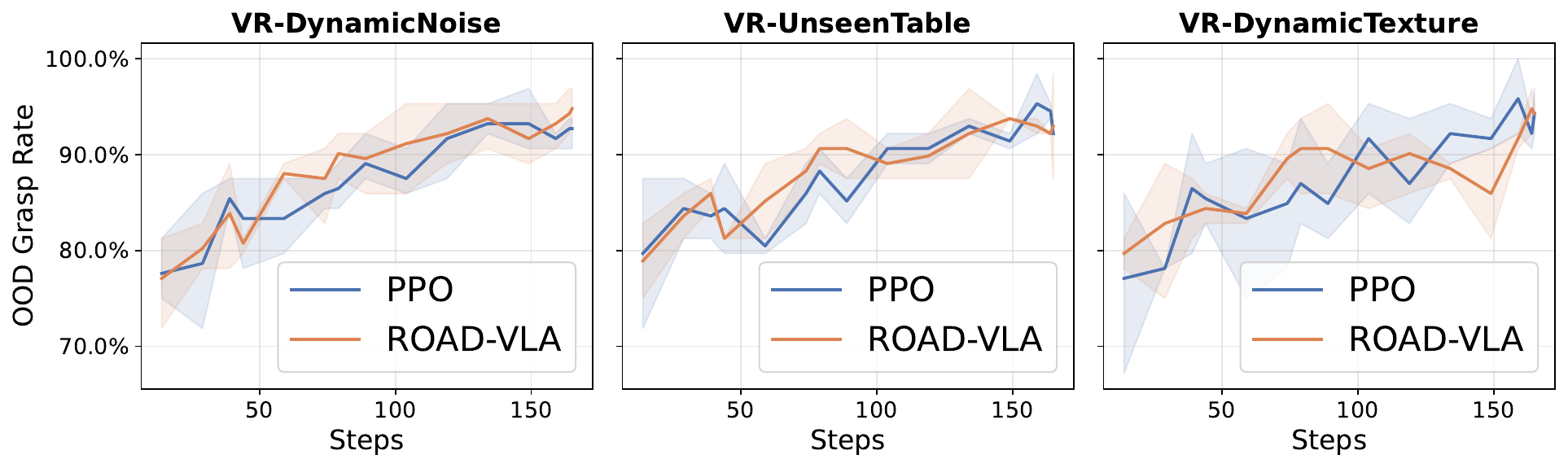}
    \caption{
    ID grasp success rate on three \texttt{VR} environments. 
    }
    \label{fig:grasp_rate_id}
\end{figure}

Refer to Figure.~\ref{fig:grasp_rate_ood} and Figure.~\ref{fig:grasp_rate_id}.

\subsection{$\alpha$ ablation} 

\begin{figure}[h]
    \centering
    \includegraphics[width=0.7\linewidth]{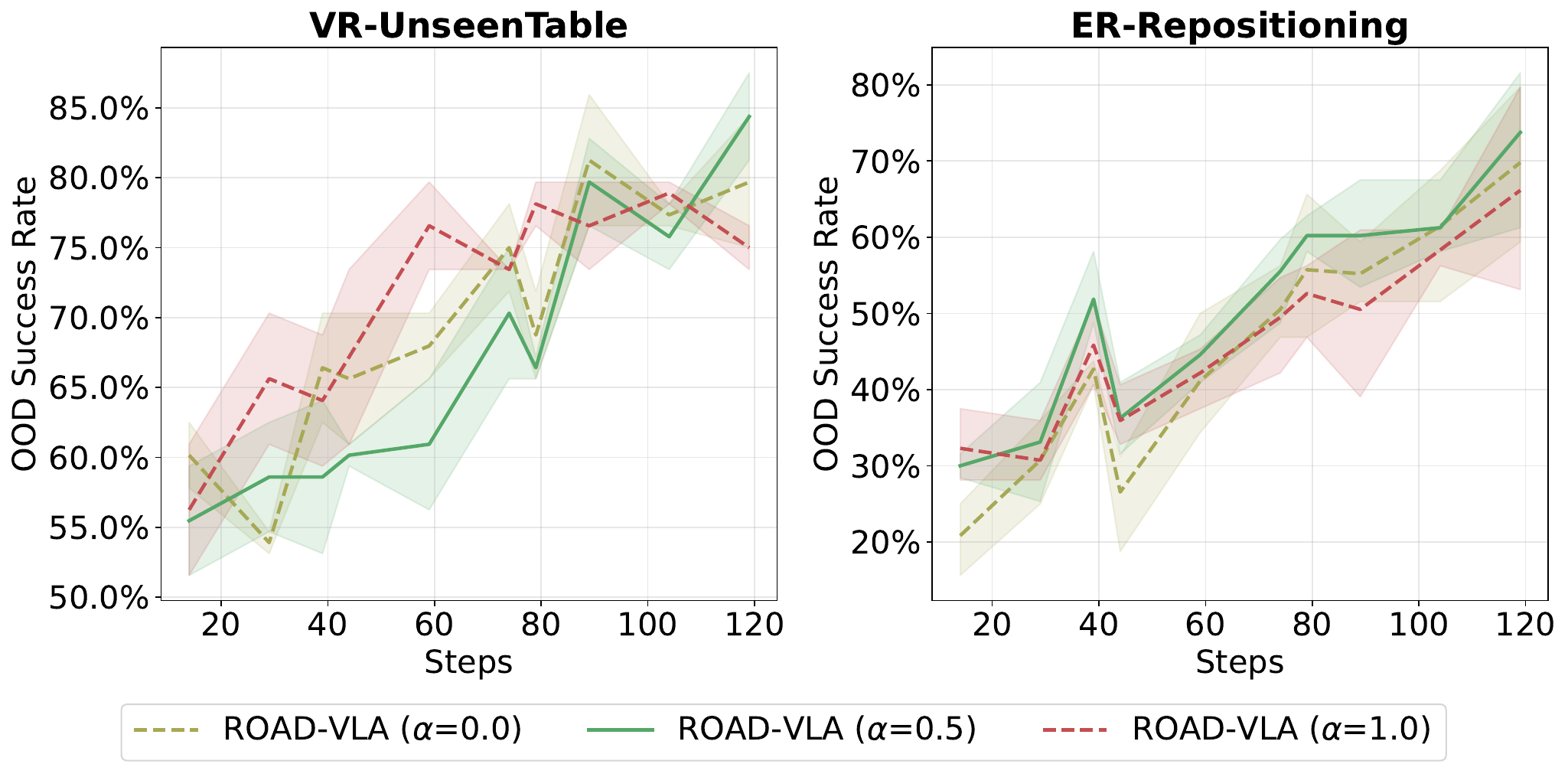}
    \caption{
    Tuning $\alpha$.
    }
    \label{fig:alpha_ablation}
\end{figure}

Refer to Figure.~\ref{fig:alpha_ablation}.

\subsection{PPO Checkpoint Sensitivity} 

\begin{figure}[h]
    \centering
    \includegraphics[width=1\linewidth]{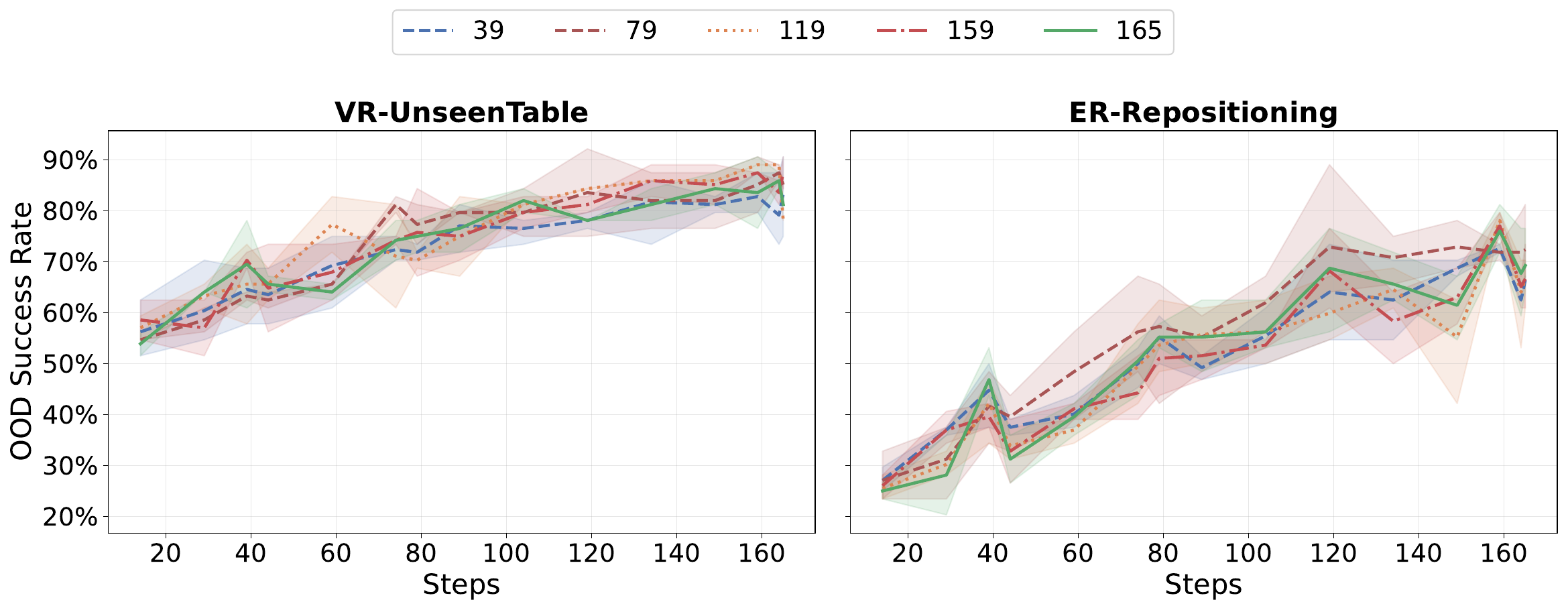}
    \caption{
    Tuning PPO warm-up checkpoint.
    }
    \label{fig:PPO warm-up checkpoint}
\end{figure}

Refer to Figure.~\ref{fig:PPO warm-up checkpoint}.

\section{Limitation and Future Work}
\label{sec:limitation}
Although ROAD-VLA improves online adaptation robustness over PPO in our evaluated settings, several limitations remain. First, our experiments are conducted on a focused set of language-conditioned manipulation tasks based on OpenVLA. While these tasks cover visual, compositional, and execution-level distribution shifts, they are still centered around pick-and-place style behaviors. Future work should evaluate ROAD-VLA on broader long-horizon, multi-stage, and multi-task manipulation settings.

Second, our evaluation is currently limited to simulation environments from the RL4VLA benchmark. Simulated perturbations allow controlled analysis of visual noise, object distractors, and execution shifts, but they may not fully capture real-world challenges such as calibration errors, lighting variation, contact dynamics, and hardware latency. Deploying ROAD-VLA on physical robots is an important next step.

Third, ROAD-VLA relies on advantage estimates and a reference PPO critic to construct the proximal teacher. Although the reference signal improves stability, its quality can affect the resulting teacher distribution. If the critic is poorly calibrated or becomes stale under large distribution shifts, the distillation target may become less reliable. Future work could explore critic-free teacher construction, uncertainty-aware advantage weighting, or adaptive mechanisms for updating the reference critic.

Fourth, the current method introduces several hyperparameters, including the distillation coefficient, logit perturbation scale, advantage clipping range, and reference critic mixing weight. While we provide a fixed configuration in our experiments, a more systematic sensitivity analysis is needed to understand how these choices affect stability and generalization. Future work could also investigate automatic tuning of the distillation strength based on policy divergence or critic agreement.

Finally, our current study focuses mainly on comparison with PPO. Additional baselines and ablations, such as text-guided distillation, uniform self-distillation, distillation without a reference critic, and different divergence objectives, would further clarify which components are most responsible for the observed gains. We believe that extending ROAD-VLA along these directions can lead to more reliable and scalable online adaptation for general-purpose VLA policies.

\clearpage

\end{document}